%% file: main.tex
\documentclass[runningheads]{llncs}

\usepackage{eccv}

\usepackage{eccvabbrv}

\usepackage{graphicx}
\usepackage{booktabs}
\usepackage{makecell}

\usepackage[accsupp]{axessibility}  %
\usepackage{xcolor} 
\usepackage{listings} %
\definecolor{commentcolor}{rgb}{0.5,0.5,0.5} %
\definecolor{keywordcolor}{rgb}{0,0,1} %
\definecolor{stringcolor}{rgb}{0.58,0,0.82} %

\usepackage{framed}

\usepackage[pagebackref,breaklinks,colorlinks]{hyperref}

\usepackage{orcidlink}

\input{preamble}

\begin{document}

\title{Taming CLIP for Fine-grained and Structured Visual Understanding of Museum Exhibits} 

\titlerunning{Taming CLIP for Visual Understanding of Museum Exhibits}

\author{Ada-Astrid Balauca\inst{1}\orcidlink{0009-0001-8466-6509} \and
Danda Pani Paudel\inst{1}\orcidlink{0000-0002-1739-1867} \and \\
Kristina Toutanova\inst{1, 3}\orcidlink{0009-0005-3458-9049}
 \and 
Luc Van Gool\inst{1,2}\orcidlink{0000-0002-3445-5711}}

\authorrunning{A.-A. Balauca et al.}

\institute{INSAIT, Sofia University, Bulgaria \\
\email{\{astrid.mocanu, danda.paudel, kristina.toutanova\}@insait.ai}
\and
ETH Zurich, Switzerland,
\email{vangool@vision.ee.ethz.ch} 
\and
Google DeepMind
}
\maketitle

\begin{abstract}
CLIP is a powerful and widely used tool for understanding images in the context of natural language descriptions to perform nuanced tasks. However, it does not offer application-specific fine-grained and structured understanding, due to its generic nature. In this work, we aim to adapt CLIP for fine-grained and structured -- in the form of tabular data -- visual understanding of museum exhibits. To facilitate such understanding we (a) collect, curate, and benchmark a dataset of 200K+ image-table pairs, and (b)  develop a method that allows predicting tabular outputs for input images. Our dataset is the first of its kind in the public domain. At the same time, the proposed method is novel in leveraging CLIP's powerful representations for fine-grained and tabular understanding. The proposed method (MUZE) learns to map CLIP's image embeddings to the tabular structure by means of a proposed transformer-based parsing network (parseNet). More specifically, parseNet enables prediction of missing attribute values while integrating context from known attribute-value pairs for an input image. We show that this leads to significant improvement in accuracy. Through exhaustive experiments, we show the effectiveness of the proposed method on fine-grained and structured understanding of museum exhibits, by achieving encouraging results in a newly established benchmark. Our dataset and source-code can be found at: \href{https://github.com/insait-institute/MUZE}{https://github.com/insait-institute/MUZE}

  \keywords{Fine-grained understanding \and Structured Data \and VLM}
\end{abstract}

\input{sections/intro}

\input{sections/related_work}

\input{sections/methods}

\input{sections/experiments}

\input{sections/conclusion}

\bibliographystyle{splncs04}
\bibliography{main}
\input{sections/appendix}

\end{document}

%% file: preamble.tex
\usepackage{multirow}

\usepackage{multicol}
\usepackage{algorithm}
\usepackage{algpseudocode}
\usepackage{lipsum}
\usepackage{pgfplots}

\usepackage{arydshln}

\usepackage{pifont}%

\newcommand{\cmark}{\ding{51}}%
\newcommand{\xmark}{\ding{55}}%

\usepackage[most]{tcolorbox}
\definecolor{block-gray}{gray}{0.95}
\newtcolorbox{myquote}{colback=block-gray,grow to right by=-0mm,grow to left by=-0mm,
boxrule=0pt,boxsep=0pt,breakable}
\makeatletter
\def\quoteparse{\@ifnextchar`{\quotex}{\singlequote}}
\def\quotex#1{\@ifnextchar`{\triplequote\@gobble}{\doublequote}}
\makeatother
\def\singlequote#1`{[StartQ]#1[EndQ]\quoteON}
\def\doublequote#1``{[StartQQ]#1[EndQQ]\quoteON}
\long\def\triplequote#1```{\begin{myquote}\parskip 1ex#1\end{myquote}\quoteON}
\def\quoteON{\catcode``=\active}
\def\quoteOFF{\catcode``=12}
\quoteON
\def`{\quoteOFF \quoteparse}
\quoteOFF

%% file: sections/intro.tex
\section{Introduction}
\label{sec:intro}

\begin{figure}
    \centering
    \includegraphics[clip, trim=2.4cm 5.5cm 2.4cm 5.5cm, width=1.0\textwidth]{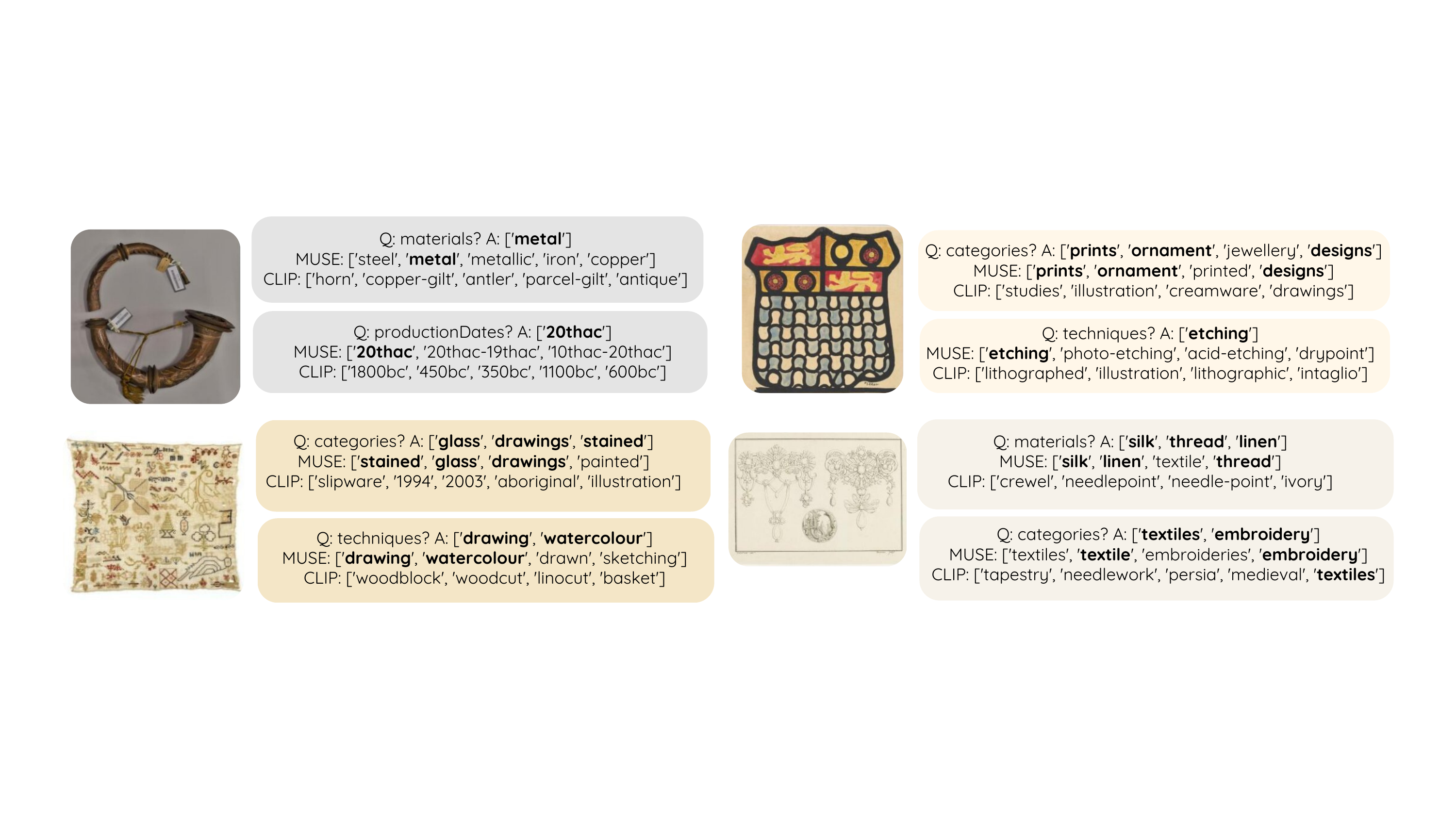}
    \caption{Fine-grained examples of \textit{materials}, \textit{categories}, \textit{techniques}, and \textit{productionDates} predicted by pretrained CLIP and the proposed method (MUZE). MUZE benefits from the tabular structure of the output as well as the context provided by the other attribute-answer pairs. }
    \label{fig:teaser}
\end{figure}

The intersection of computer vision and natural language processing has led to the development of vision-language models (VLMs) such as CLIP~\cite{radford2021learning}, ALIGN~\cite{jia2021scaling}, and others~\cite{kim2021vilt,singh2022flava,bao2021beit}.
This has significantly improved our ability to understand visual content within the context of natural language descriptions.
CLIP's robust architecture and widespread adoption have proven its effectiveness in bridging the gap between textual annotations and visual data, making it useful for a wide range of nuanced tasks~\cite{wei2023improving,wang2022medclip,cui2024learning, zhai2022lit, pham2023combined,rasheed2023fine}. In fact, visual understanding by vision-language alignment pre-training is a scalable technique, which is also evident by the remarkable success of the aforementioned methods in diverse domains.

In general, the pre-trained VLMs --- which offer rich visual representations ---  are developed to be adapted separately to the downstream vision tasks\footnote{Some other variants, such as~\cite{alayrac2022flamingo} also exist, however,  a separate vision-only representation models are often preferred primarily due to the development ease. }.
When the downstream tasks are still the vision-language type, it is relatively easy to make such adaptations even for different domains. However, doing the same for the tasks with different input/output modalities is not trivial. Therefore, several methods have been developed recently to address the problem of different input and outputs types. Some such examples include object detection~\cite{gu2021open,bangalath2022bridging,zhou2022detecting}, semantic segmentation~\cite{ding2021decoupling,Boyi2022,zhou2022extract}, video understanding~\cite{ju2022prompting,ni2022expanding,wang2021actionclip,rasheed2023fine}, and others~\cite{lin2024fine,meng2023foundation}.

We aim to use pre-trained Vision-Language Models (VLMs) to understand museum exhibits visually. A few examples of such understanding are shown in Figure~\ref{fig:teaser}. Museums hold a vast collection of cultural heritage and historical artifacts that belong to different epochs, civilizations, and geographies. The exhibits are well-documented, and their data is usually structured in tables that list attributes such as age, origin, material, and cultural significance~\cite{becattini2023viscounth,marty2008museum,nishanbaev2019survey}. This structured approach is essential as it reflects how museum collections are organized, where exhibits share similar attributes. The table representation helps human experts categorize and understand artifacts better. For example, understanding the material and origin of a vase can significantly inform its historical context and cultural value. However, converting the rich visual information of museum exhibits into structured data is a unique challenge. The descriptions and labels associated with the exhibits require a method that can capture the nuanced interplay between visual cues and textual information to produce structured data. To the best of our knowledge, there is no such VLM-based method which also benefits from the large-scale pre-training.

\begin{figure}
    \centering
    \includegraphics[clip, trim=4.8cm 3cm 4.8cm 3cm, width=1.0\textwidth]{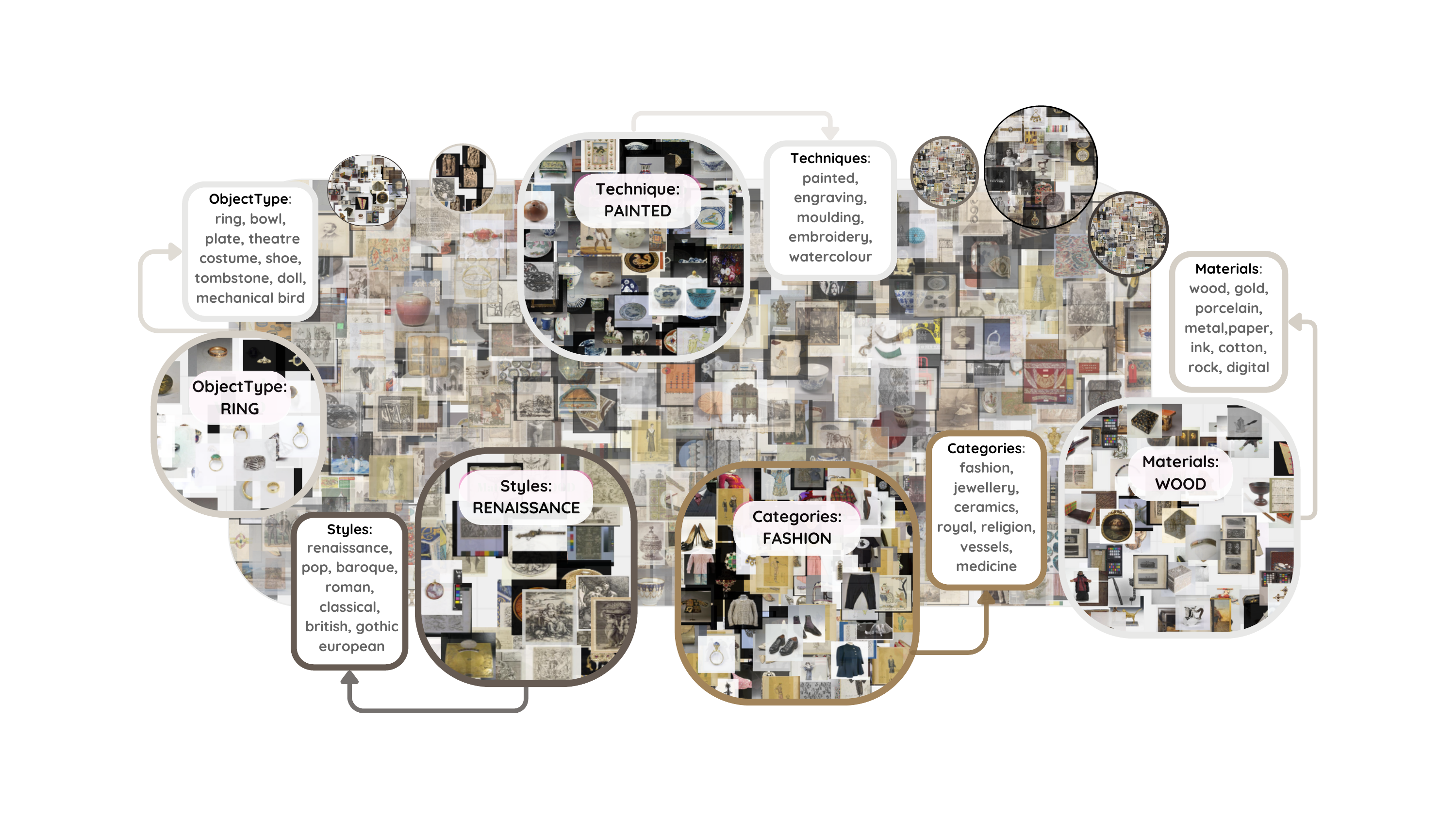}
    \caption{Our dataset  contains a variety of attributes with annotation of corresponding fine-grained labels.  Some of those attributes and a subset of their possible values are highlighted here, along with sample images corresponding to a chosen class.  }
    \label{fig:teaser2}
\end{figure}

This paper first introduces a dataset of images and their corresponding tabular descriptions, which we then use to benchmark CLIP-based models in understanding museum exhibits. Later, a novel and effective method is proposed to predict structured outputs, in a tabular form,  from input images. On the dataset side, we have collected and curated a comprehensive dataset comprising over 200K image-table pairs. A visual summary of our dataset is shown in \cref{fig:teaser2}. 
On the method side, we propose a simple yet effective method, referred to as MUZE~\footnote{We refer to both the dataset and method with the same name. In case of ambiguity we use MUZE (dataset) or MUZE (method).}, which benefits from CLIP's large-scale pre-training and  provides the desired tabular outputs. MUZE is built around a transformer-based parsing network called parseNet. It processes CLIP's image embeddings and attribute queries to produce precise tabular entries. This method enhances CLIP's utility for the specific application and introduces a mechanism for leveraging contextual information within tabular data to improve accuracy in generating entries for missing attributes.

Our work extends the application range of vision-language models like CLIP, and also sets a new benchmark for fine-grained, structured visual understanding in the museum domain. By making our dataset and source code publicly available, we aim to foster further research and application development in this area. Through exhaustive experiments, we analyze the performance of our approach and showcase its effectiveness,  highlighting its potential to transform the way we interact with and understand visual information in complex, information-rich environments like museums.

The major contributions of this paper are threefold:
\begin{itemize}
\item \emph{Dataset and benchmark baseline:} We introduce a dataset of 200K+ image-table pairs for museum exhibits, suitable to build vision-language models for structured prediction and establish a CLIP-based benchmark for it. 

\item \emph{Method:} We develop a novel method using CLIP and a transformer-based parsing network (parseNet) for precise image-to-table data mapping.

\item \emph{Results and insights:} Our method provides state-of-the-art results for the task at hand. Furthermore, we provide several insights about the collected dataset as well as our method in leveraging the available context. 
\end{itemize}

%% file: sections/related_work.tex
\section{Related Work}
\label{sec:related work}

\noindent \textbf{Vision-Language Pre-training Models}: 
Training models easily adaptable to other tasks has become increasingly useful. Pre-trained models like CLIP~\cite{radford2021learning} and ALIGN~\cite{jia2021scaling}, are used in both unimodal and multimodal tasks covering a wide variety of applications in zero-shot recognition \cite{zhang2021tip,zhou2022learning,zhou2022conditional}, object detection \cite{gu2021open,bangalath2022bridging,zhou2022detecting}, image segmentation \cite{ding2021decoupling,li2022languagedriven,zhou2022extract}, and more \cite{conde2021clip,chen2020uniter,kamath2021mdetr,li2019visualbert,li2020oscar,maaz2022class,lu2019vilbert}. As the pre-trained models already have built a deep knowledge about general concepts, they may be useful tools in understanding specific domains such as cultural heritage and museums.

\begin{table*}
\vspace{-0.5cm}
\centering    
\caption{Comparison between our dataset and related datasets from the literature. We compare the data domains, their size and if their structure. We are interested in images as well as captions or tabular data related to images. Note that not many datasets regarding Cultural Heritage also include tabular data. }
\label{tab:dataset_comp}
\begin{tabular}{@{}lccrccc@{}}
\toprule
Dataset&Heritage&Domain&\#images&Captions&Tabular& Public\\
\hline
\textsc{HybridQA}~\cite{chen2020hybridqa}&\xmark&-&-&\cmark&\cmark&\cmark\\
\textsc{ManyModalQA}~\cite{hannan2020manymodalqa}&\xmark&-&3K&\cmark&\cmark&\cmark\\
\textsc{MultiModalQA}~\cite{talmor2021multimodalqa}&\xmark&-&57K&\cmark&\cmark&\cmark\\
Sheng et al.~\cite{sheng2016dataset}&\cmark&Archaeology&160&\cmark&\xmark&\xmark \\
AQUA~\cite{garcia2020dataset}&\cmark&Art&21K&\cmark&\xmark&\cmark\\
iMet~\cite{zhang2019imet}&\cmark&Art\&History&155K&\cmark&\xmark&\cmark \\
VISCOUNTH~\cite{becattini2023viscounth}&\cmark&Art&500K&\xmark&\cmark&\xmark \\
MUZE (Ours)&\cmark&Art\&History&210K&\cmark&\cmark&\cmark \\
\bottomrule
\end{tabular}
\vspace{-0.3cm}
\end{table*}

\begin{table*}
\centering    
\caption{Comparison between our method and different related methods from the literature. We are interested in the method's multi-modality and its expression. All the analyzed methods rely on a separation between visual and textual input instead of feeding all the information to a single VLM entity. We also observe a lack of methods leveraging tabular structured input. }
\label{tab:method_comp}
\begin{tabular}{@{}lcccc@{}}
\toprule
Method&Use VLM&Multi-modal&UseContext&TabularInput\\
\hline
\textsc{ManyModalQA}~\cite{hannan2020manymodalqa}&\xmark&\cmark&\xmark&\xmark \\
ImplicitDecomp~\cite{talmor2021multimodalqa}&\xmark&\cmark&\cmark&\xmark \\
\textsc{HybridQA}~\cite{chen2020hybridqa}&\xmark&\xmark&\cmark&\xmark \\
VISCOUNTH~\cite{becattini2023viscounth}&\xmark&\cmark&\cmark&\xmark \\
Bai et al.~\cite{bai2021explain}&\xmark&\cmark&\xmark&\xmark \\
VIKINGfull~\cite{garcia2020dataset}&\xmark&\cmark&\xmark&\xmark \\
MUZE (Ours)&\cmark&\cmark&\cmark&\cmark \\
\bottomrule
\end{tabular}
\vspace{-0.3cm}
\end{table*}

\noindent  \textbf{Structured Data Generation}: 
Generating structured data corresponding to images has been explored in visual document domains, such as scanned receipts, where the task is to predict attribute-value pairs corresponding to form-like input documents~\cite{formNet,donut,hwang2019post} or underlying tabular data corresponding to figures and graphs~\cite{deplot,siegel2016figureseer,luo2021chartocr,masry2022chartqa}.  In these works, elements of the structured output often correspond to individual texts in the image and the task is solved by labeling image regions. Pre-trained VLM models used in these approaches are specialized to visual document domains, although some recent pre-trained models have shown strong capabilities on natural images as well~\cite{pix2struct, pali}. 

In this paper, we formulate a structured data prediction task where the attributes often represent global properties of the image and have joint dependencies across attribute values.

\noindent  \textbf{Digital Humanities and Cultural Heritage}: 
For qualitative visual understanding we need to identify and to extract not only informative images but also many reliable textual information. However, achieving useful texts requires expertise in a specialized domain, which is expensive and time-consuming, being a limiting factor in the collection of data \cite{dataset2011novel,maji2013fine,wah2011caltech,sheng2016dataset}. 

Previous attempts on leveraging Cultural Heritage data approached it from a multi-modal perspective, \cite{hannan2020manymodalqa,talmor2021multimodalqa,becattini2023viscounth,bai2021explain,garcia2020dataset,lu2022data,gao2015you} usually without using VLMs to solve the task and without exploiting the benefits provided by the tabular format for explaining exhibits' historical importance. (see \cref{tab:dataset_comp,tab:method_comp}).

%% file: sections/methods.tex
\section{The MUZE Dataset}

We collected 210K image-table pairs by gathering data scraped from the public domains on the internet. We observed the most common structure of the provided descriptions of exhibits to be in tabular format, having multiple columns representing attributes. We found this format helpful for providing context to each exhibit image. The images we collected were captured by museum professionals, and the labels authored by curators are separated into multiple attributes forming the tables we use for image understanding.

\subsection{Data Curation}
We engaged with museum experts across multiple institutions, collaborating and learning from them to enhance our understanding of museum practices in order to improve our data curation skills, aligning our perspective with domain experts.
We curated the data in several stages: %
\begin{enumerate}
    \item \textbf{Elimination of the unnecessary} punctuation, the letters that do not belong to the Latin alphabet, redundant terms (as ``representation'', ``made in'', ``note'', ``translation'') and terms that express uncertainty (as ``probably'', ``possibly'', ``about'', ``around'', ``perhaps''). 
    \item \textbf{Shortening the text} of the attributes/columns that contain considerable amount of text. We separate each paragraph into phrases and extract 10 keywords with maximum n-gram size of 3 from each phrase using \texttt{yake}\cite{campos2020yake}, which takes into account frequency and POS tagging. 
    \item \textbf{Temporal references} (such as productionDates) needed particular attention. For them, we eliminated terms not related to temporal information, vaguely expressed dates, or different notation systems, transforming all years in the same century related format, keeping intervals and suffixing each century by AC or BC, for consistency (e.g. 17th AC, 5th BC-4th AC).
\end{enumerate}

\subsection{MUZE Benchmark}

We build the dataset as being divided into \textit{train}, \textit{val} and \textit{test}, having about \textit{178k}, \textit{16k} and \textit{33k} samples respectively. The data is provided as csv files, with one column for each attribute, having extra columns for object id, image path, caption. To reduce the size of the csv files, we replaced the value of each attribute/column with a list of numerical ids. The mapping between the set of values of each attribute/column and the numerical ids is provided as a separate json file.

For building the captions necessary to test and fine-tune CLIP we concatenate the values of the alphabetically ordered attributes/columns of interest, as the data was curated carefully and the use of the keywords helped us to concentrate the meaningful information in a usable way.

As benchmark methods we used CLIP pre-trained, along with two variants called CLIP-FC and CLIP-FA, which we fine-tuned on captions and attributes respectively. For more implementation details see \cref{subsec:implementtaion_details}. On average, CLIP-FA obtains better results than CLIP-FC and CLIP.

\begin{figure}
    \centering
    {\includegraphics[clip, trim=3cm 6.8cm 3.1cm 8.1cm, width=0.23\textwidth]{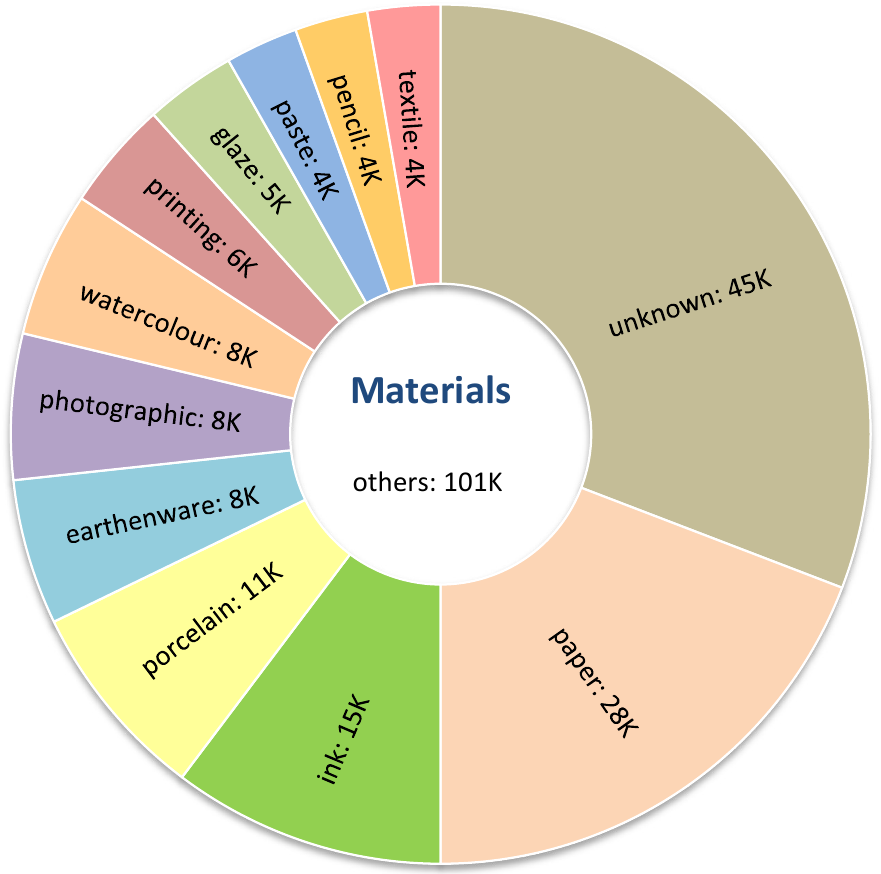}}
    {\includegraphics[clip, trim=3.5cm 6.5cm 2.2cm 8.0cm, width=0.23\textwidth]{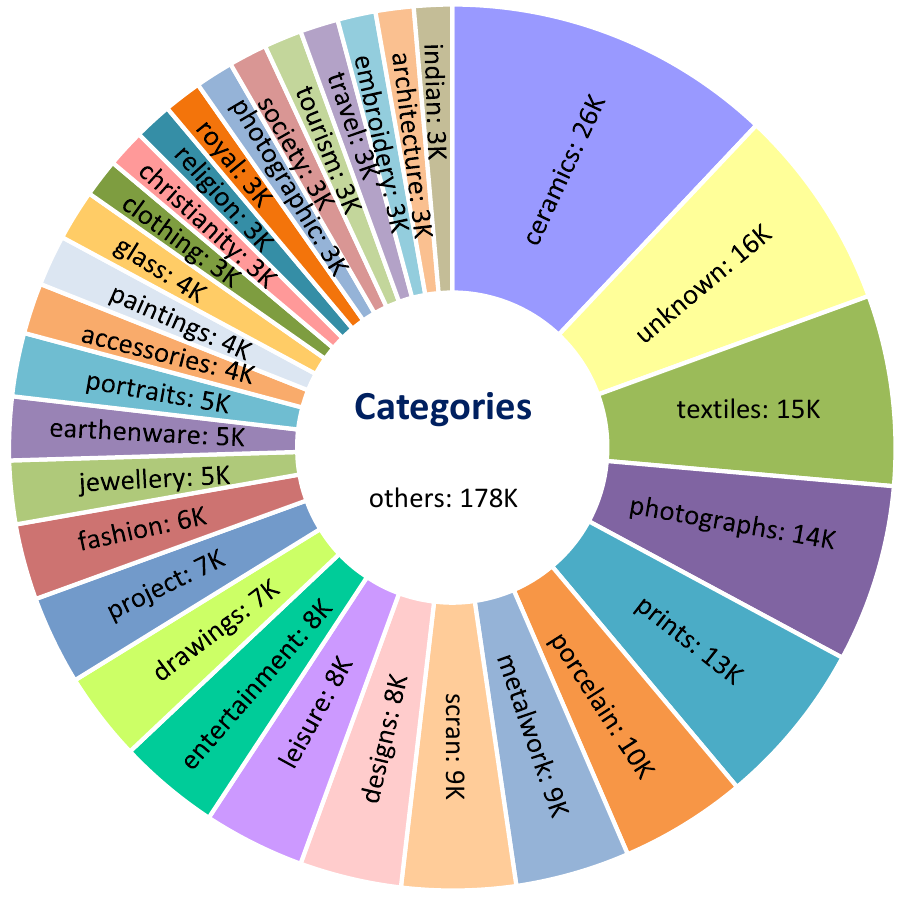}}
    {\includegraphics[clip, trim=3cm 6.8cm 3.1cm 8.1cm, width=0.23\textwidth]{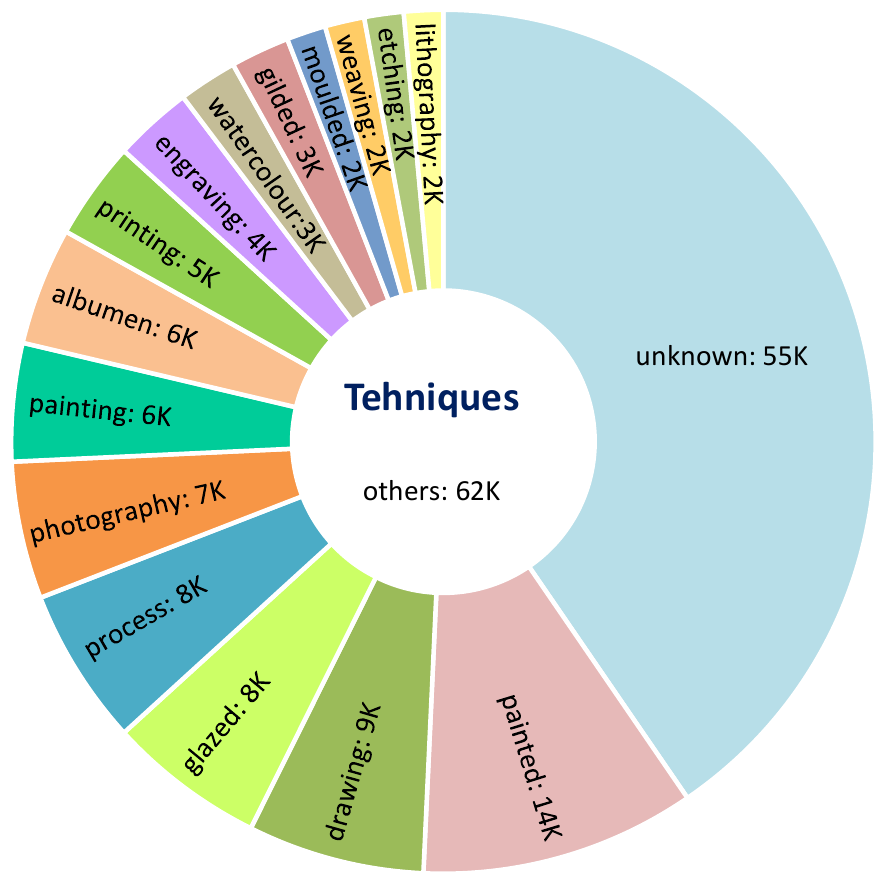}}
    {\includegraphics[clip, trim=2.5cm 6.2cm 2.5cm 7.7cm, width=0.23\textwidth]{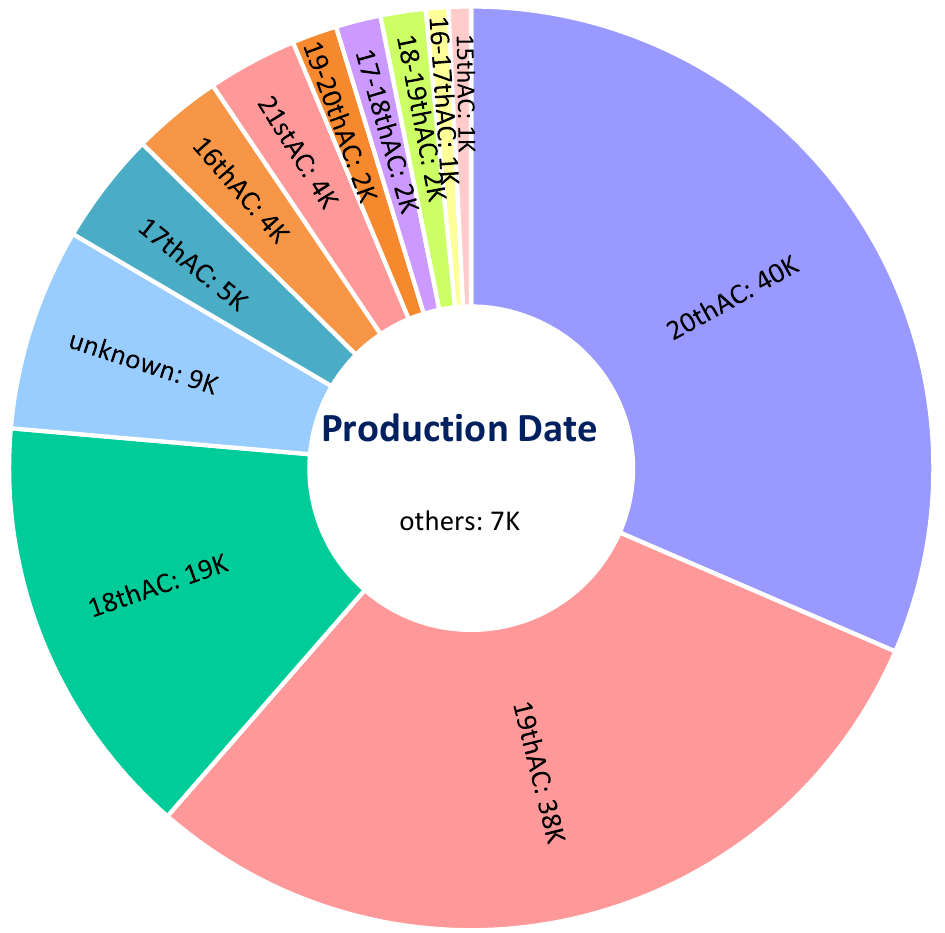}}
    \caption{Quantitative analysis of the value distribution for some attributes considered suitable for classification. Each chart displays the number of images which share the most common values for the corresponding attribute. The values with counts lower than a threshold were cumulated inside the chart under the name of \textit{others}. Note that different values for a given attribute don't necessarily describe disjoint sets, e.g. some objects can have both \textit{paper} and \textit{ink} as values for the \textit{materials} attribute.}
    \label{fig:attributes_dist}
\end{figure}

\subsection{Data Exploration}
\label{sec:data_exploration}
The MUZE dataset was collected from different sources. During the analysis, we separate the dataset in two parts, A-MUZE and B-MUZE, based on the different configuration of the attributes available for the images. The A-MUZE subset contains tables with 18 attributes, such as ``artistMakerPerson''
and ``historicalContext'', while B-MUZE has 12 attributes, such as ``Producer Name'', ``Production Date'', and ``Object Type''.
Most images have a dimension similar to or slightly larger than the standard input of CLIP (224$\times$224 pixels). The images are depicting objects related to art, human history, archaeology, and culture.

We considered the attributes which have between 200 and 10.000 values, like ``categories'', ``productionDates'', and ``materials'',  as classifiable, and the rest as textual. Using the pie charts in \cref{fig:attributes_dist}, we present the distribution of the values for several classifiable attributes, noting that values for some attributes are unbalanced.

\begin{figure}
    \centering
    \begin{subfigure}{0.45\linewidth} %
        {\includegraphics[clip, trim=7.7cm 2cm 7.5cm 2cm, width=1.0\textwidth]   {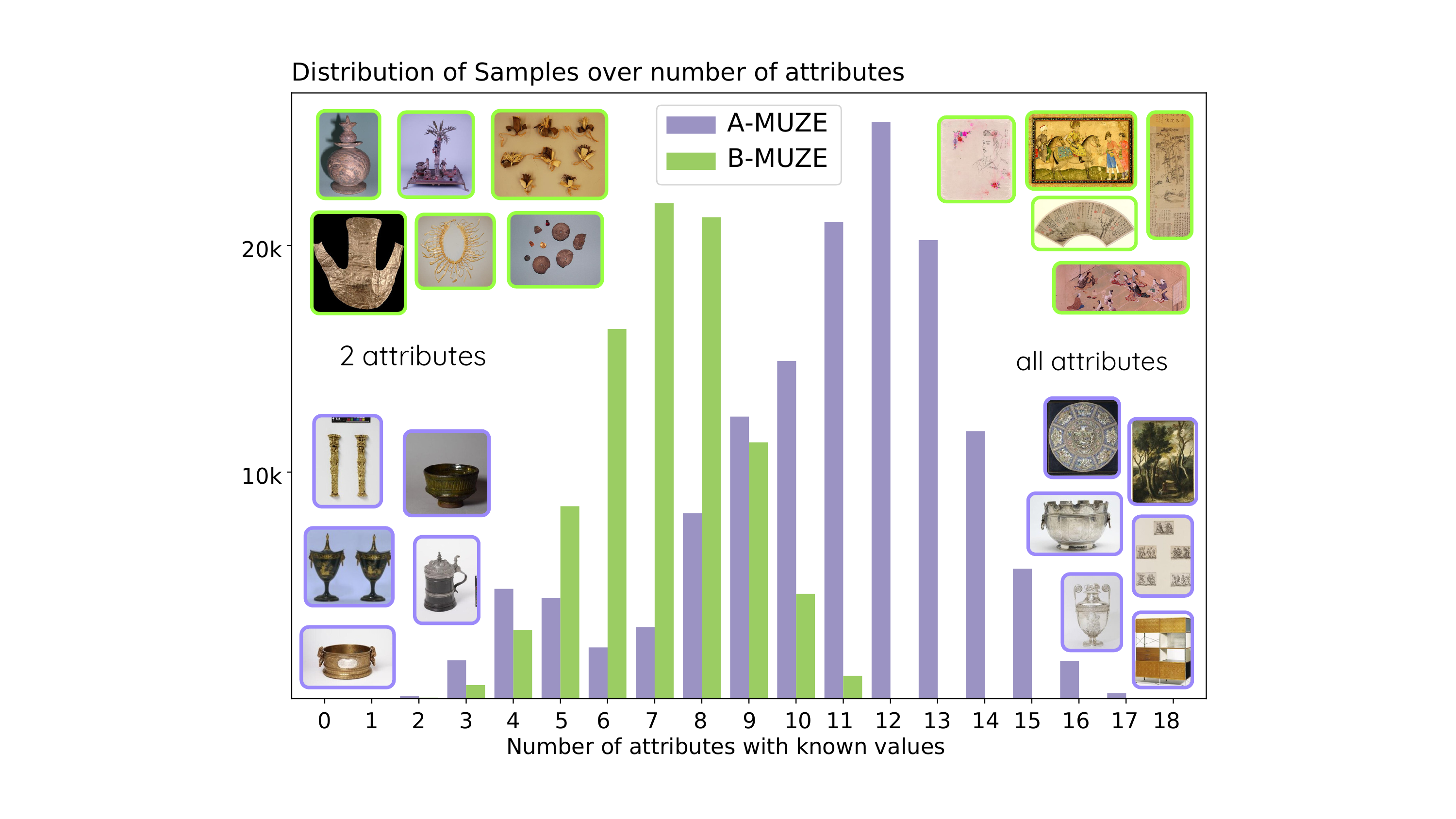}}
        
    \end{subfigure}
    \begin{subfigure}{0.54\linewidth} %
        {\includegraphics[clip, trim=10.5cm 9.9cm 10.5cm 1.5cm, width=1.0\textwidth]{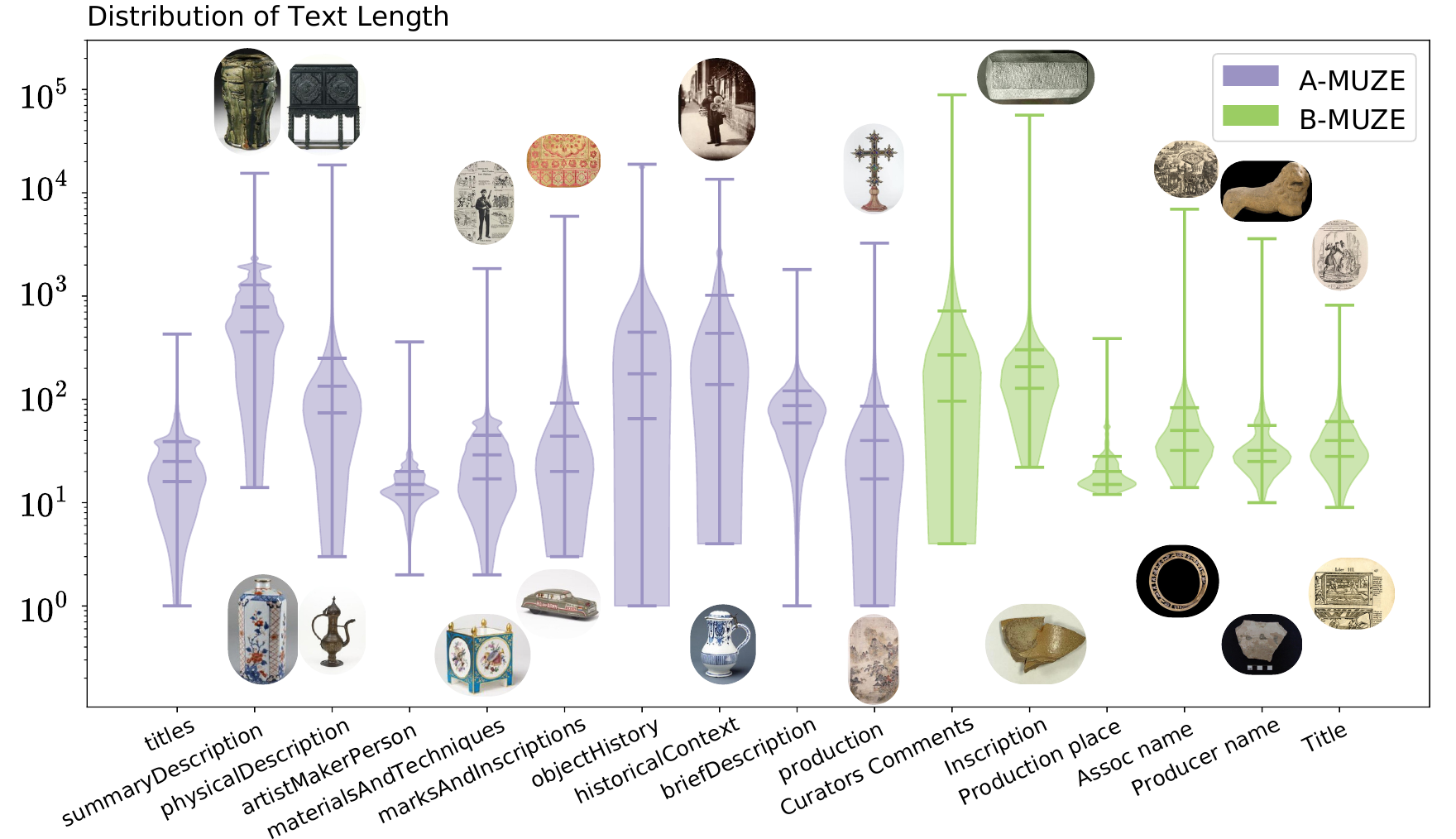}}
        
    \end{subfigure}
    \vspace{-0.3cm}
    \caption{Quantitative analysis of various collected attributes across samples. \textit{Left:} Histogram of per-sample count of non-empty attribute columns for the two sub-datasets; A-MUZE has a total of 18 attribute columns, while B-MUZE has 12. We also illustrate samples of images with few (2) or many (18 or 12) non-empty attributes. \textit{Right:} Violin plots representing the distribution of text lengths (number of characters) for textual attributes.}
    \label{fig:data_statistics}
\end{figure}

In \cref{fig:data_statistics} \textit{left} we show the distribution of the samples over the number of attributes, targeting the understanding of the amount of context available for each instance. We present in the upper part examples of B-MUZE images and in the lower part examples from A-MUZE. On the left side, the images have the lowest number of known attributes, while on the right side we show images for which all attributes have known values. We remark the available context for the objects follows a normal distribution on both subsets of the dataset (we note again that the A-MUZE dataset has 18 attributes for each exhibit, while the B-MUZE dataset has 12 attributes). We observe that in both data subsets, for most exhibits more than half of the attributes are known.

In \cref{fig:data_statistics} \textit{right} we analyze the length distribution for the textual attributes. Most of these attributes have a high variance, their lengths spanning across 3-4 orders of magnitude. We display examples for shortest length values in the lower part, and for the longest length values in the upper part of each violin plot, for some of the attributes. We observe there is a large variety of objects regardless of textual description length, but there exist a tendency for shorter descriptions on incomplete, small or very simple objects, while more complex or older objects have longer descriptions.

\section{The MUZE Method}
\begin{figure}
    \centering
    {\includegraphics[clip, trim=1.5cm 3.5cm 1.4cm 3.5cm, width=1.0\textwidth]    {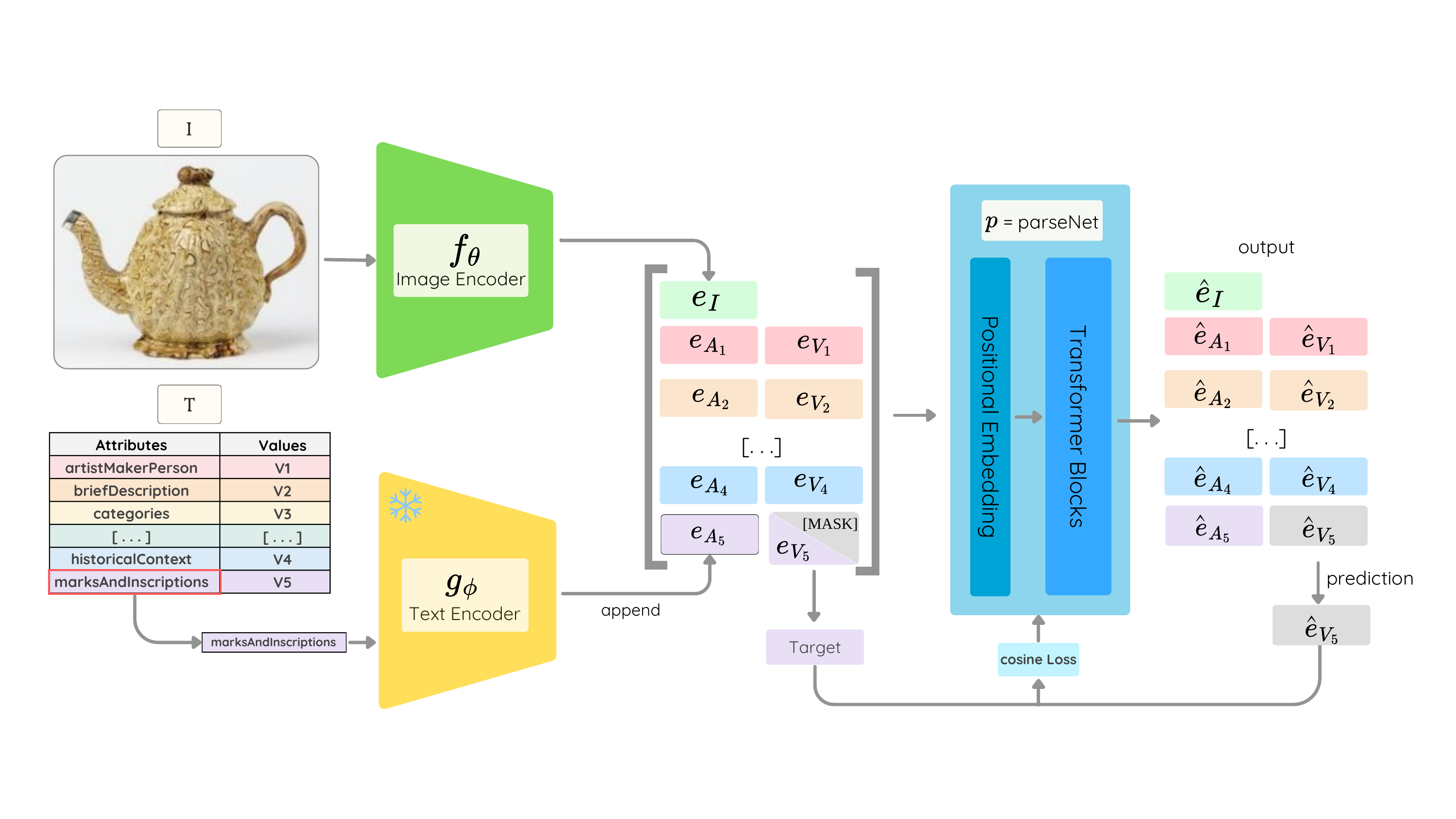}}
    \caption{Schematic representation of our proposed method (MUZE). We show the process of obtaining CLIP embeddings for the input image ($\mathsf{e}_I$), attribute names ($\mathsf{e}_{A_i}$) and attribute values ($\mathsf{e}_{V_i}$). After replacing the embeddings of the query attribute values with \texttt{[MASK]} tokens we pass the obtained sequence of embeddings through parseNet to obtain the predicted embeddings for the query attributes. The CLIP Image Encoder and parseNet are trained to maximize the cosine similarity between the target and predicted embeddings.}
    \label{fig:method}
\end{figure}

We use the CLIP's encoder: $\mathsf{e}_I=f_{\theta}(\mathcal{I})$ that maps the input image $\mathcal{I}$ to its visual embedding $\mathsf{e}_I$ with the help of network parameters $\theta$. Similarly, the text encoder of CLIP $g_\phi(.)$ maps the input text to its embedding, using the frozen network parameters $\phi$. Let $(A,V)$ be the tuple of attribute and its value for the image $\mathcal{I}$.  Then, we define a tabular data as a set of such tuples: $\mathcal{T} = \{(A_i,V_i)\}_{i=1}^n$, without the loss of generality. Now we are interested to process the tabular data $\mathcal{T}$ using the text encoder $g_\phi(.)$. Let $\mathcal{E_T} = \{(\mathsf{e}_{A_i},\mathsf{e}_{V_i})\}_{i=1}^n$ be the a set of tuples of embeddings obtained for $\mathcal{T}$ such that $\mathsf{e}_{A_i} = g_\phi(A_i)$ and $\mathsf{e}_{V_i} = g_\phi(V_i)$. For simplicity, we represent $\mathcal{E_T} = G_\phi(\mathcal{T})$ for text encoding part where the tabular data $\mathcal{T}$ is encoded to its corresponding tabulated embeddings $\mathcal{E_T}$.  
Our choice of taming CLIP is motivated from the fact that methods which map image+text to joint embedding (e.g. BLIP~\cite{li2022blip}) are not trivial to exploit for the tabular query-answer settings in an efficient manner. This is because, the same attributes are shared across multiple images and values/answers. Exploiting this structure efficiently requires attributes to be embedded separately.

During the inference, we wish to predict the missing values of some attributes. Therefore, we mask some attributes' values, by dividing the table into query $\mathcal{T}_q$ and context $\mathcal{T}_c$ such that $\mathcal{T} = \mathcal{T}_q \cup \mathcal{T}_c$  and $ \mathcal{T}_q \cap \mathcal{T}_c = \emptyset$. Now, we denote the tuple's embedding sets $\mathcal{E}_{\mathcal{T}_q}$ and $\mathcal{E}_{\mathcal{T}_c}$ respectively for the query and context tables.  We further define remaining sets, $\mathcal{A}_q = \{A|A\in\mathcal{T}_q\}$ and $\mathcal{V}_q = \{V|V\in\mathcal{T}_q\}$, and their corresponding embedding sets  $\mathcal{E}_{A_q}$ and $\mathcal{E}_{V_q}$, respectively. Now, we are interested to learning the following mapping function, 
\begin{equation}
    \mathcal{V}_q := H(\mathcal{I}, \mathcal{A}_q,\mathcal{T}_c ).
\end{equation}
In other words, we are interested to perform the visual understanding of the image $\mathcal{I}$ by generating the values $\mathcal{V}_q$ corresponding to the queries $\mathcal{A}_q$, in the presence of the context $\mathcal{T}_c$. We realize such mapping with the help of the text and image encoders, together with the proposed network, i.e. parseNet. Our parseNet $p_\psi(.)$, parameterized by $\psi$ processes the embeddings of image, query attributes, and context to predict the missing query values, which is given by,

\begin{equation}
    \hat{\mathcal{E}}_{V_q} := P_\psi(\mathsf{e}_I, \mathcal{E}_{A_q},\mathcal{E}_{\mathcal{T}_c}).
\end{equation}

The different components of our method, along with some helpful notations, are shown in \cref{fig:method}. In the following, we first present the design of  parseNet, and then its training loss and protocols. 

\subsection{The Parsing Network (parseNet)}
The architecture of the parseNet is based on the same Transformer architecture used by CLIP's Text Encoder. The base configuration used for parseNet is a 2-layer 512-wide Multi-Head Attention Transformer with 8 attention heads.

The input for parseNet is a concatenation of the embedding vectors $\mathsf{e}_I, \mathsf{e}_{A_i}\in\mathcal{E}_{A_q}$ and $ \mathsf{e}_{T_j}\in\mathcal{E}_{\mathcal{T}_c}$ generated by passing the input through CLIP's Image and Text Encoders respectively. In order to predict the embedding of the missing attribute values $\hat{\mathsf{e}}_{V_i}\in\hat{\mathcal{E}}_{V_q}$ we insert \texttt{[MASK]} tokens after the embedding of each query attribute $\mathsf{e}_{A_i}$. We collect the predicted embeddings from the corresponding token positions in the output of the transformer, as illustrated in \cref{fig:method}.
\subsection{Training Loss and MUZE Algorithm}

For training parseNet and the Image encoder, we used the cosine loss \cite{barz2020deep}, in order to enforce similarity between the predicted and target embeddings. In the case where the query attribute has multiple associated values, the target embedding is computed as the sum of the embeddings of each of the attribute's values.
\begin{equation}
    \mathcal{L} = \sum\limits_{{\mathsf{e}}_{V_q}\in\mathcal{E}_{V_q}}(1 - \cos(\hat{\mathsf{e}}_{V_q}, \mathsf{e}_{V_q}))
\end{equation}

\label{sec:method}

In the following, we highlight the outline of the MUZE Algorithm.

\begin{minipage}[c]{0.99\textwidth}
\begin{framed}
\begin{lstlisting}[mathescape=true,language=Python, caption=Step-by-step processing of the MUZE algorithm.]

image, table  = init(data) # obatain image $\mathcal{I}$ and  table $\mathcal{T}$
    
for image, table in data:
    e_img = embedImage(image) # $\mathsf{e}_I$ from image encoder $f_\theta(\mathcal{I})$
    T_query, T_contex = divideTable(table) # divide query and context
    for attribute,value in T_context:
        # $\mathsf{e}_{A_i}$ and $\mathsf{e}_{V_i}$ from image encoder $g_\phi(\mathcal{T}_c)$
        context_emb.add(embedText(attribute),embedText(value)) 
    for attribute,value in T_query:
        # $\mathsf{e}_{A_i}$ and [MASK] from image encoder $g_\phi(\mathcal{T}_q)$
        attribute_query_emb.add(embedText(attribute),[MASK]) 
        tgt_emb.add(embedText(value))

    # Passing through the parseNet Transformer 
    pred_emb=parseNet(e_img,attribute_query_emb,context_emb)
    
    # Calculate error and optimize
    loss = calc_loss(tgt_emb, pred_emb)
    optimize(loss)
\end{lstlisting}
\end{framed}
\end{minipage}

%% file: sections/experiments.tex
\section{Experiments}
\label{sec:experiments}

\subsection{Attribute Prediction Task and Metrics}
We formalize our task of predicting the value for a given query attribute $A_q$ as retrieving the most suitable answers from the list of all possible values of the query attribute. For the MUZE method we retrieve $\hat{\mathsf{e}}_{V_i}$ as described in \cref{sec:method}, while for CLIP and its fine-tuned variants we consider $\hat{\mathsf{e}}_{V_i} = \mathsf{e}_I = f_\theta(\mathcal{I})$. We then compute the cosine similarity between $\hat{\mathsf{e}}_{V_i}$ and the embedding $\mathsf{e}_{V_i} = g_\phi(V_i)$ of each value $V_i$ from the set $\mathcal{S}_V(A_q)$ of possible values for the attribute $A_q$, to obtain an ordering of the values. In the case of textual attributes, this approach will become similar to simple caption retrieval task, without requiring any changes.

Since the task can be seen as a multi-class multi-label classification, we use multiple metrics to evaluate the performance of our methods:
\begin{enumerate}
    \item \textbf{Mean Average Precision across Classes}: Since most of the attributes have very unbalanced distributions for their values (see \cref{fig:attributes_dist}) we compute the average precision over each class in $\mathcal{S}_V(A_q)$ and report the mean. 
    \item \textbf{Mean Average Accuracy across Samples}: Computes the percentage of the predictions that are part of the set of correct answers, averaged across the number of total predictions taken into account.
    \item \textbf{Accuracy Top 1}: Fraction of input samples for which the top prediction is part of the set of correct answers.
    \item \textbf{Hit Rate Top 5}: Fraction of input samples for which at least one of the top 5 predictions is part of the set of correct answers.
\end{enumerate}

\subsection{Implementation Details}
\label{subsec:implementtaion_details}

In all our experiments we use the openCLIP \cite{cherti2023reproducible} implementation of the CLIP model, with VIT-B-32 architecture, with the \texttt{laion2b\_s34b\_b79k} pre-trained weights. For further details, as well as our code and dataset please refer to: \href{https://github.com/insait-institute/MUZE}{https://github.com/insait-institute/MUZE}

\subsubsection{Fine-tuning CLIP}
We fine-tune for 100 iterations with a batch size of 1024, learning rate 1e-4 and a cosine annealing learning rate schedule with 100 warm-up steps and the AdamW optimizer \cite{loshchilov2017decoupled}, 
with weight decay 0.1. We freeze the weights of the text encoder and the first 8 layers of the image encoder.

For a better analysis of CLIP's capabilities, we choose to fine-tune and evaluate it in two different settings: 
\begin{enumerate}
    \item \textbf{CLIP-FC}: for each image we use the full caption obtained by concatenating all values from all attributes.
    \item \textbf{CLIP-FA}: for each attribute we fine-tune a different image encoder by using the same settings, the only difference being that instead of caption we use the concatenation of the values of the corresponding attribute.
\end{enumerate}
We evaluate both methods on the attribute prediction task.

\subsubsection{Training parseNet}
We use mostly the same hyperparameters as for fine-tuning CLIP, except for the warm-up which we set to 0 because we start from a randomly initialized network. We train parseNet and the unlocked layers of the image encoder as described in \cref{sec:method}. As in the case of \textbf{CLIP-FA}, we train a different model for predicting each attribute in the dataset. Depending on the backbone used, we named our method \textbf{MUZE-C}, \textbf{MUZE-CFC}, or \textbf{MUZE-CFA}.

\subsubsection{Hardware} We train and evaluate our models using 8$\times$NVIDIA L4 GPUs.

\subsection{Main Results}
In \Cref{tab:vamus,tab:britmus} we present the results of our MUZE-CFA method compared with CLIP based baselines on both A-MUSE and B-MUZE.
We observed a tendency for CLIP-FA to have much better results than CLIP-FC. Our model achieves about 1.5-2 times better average results than the best CLIP fine-tuned on any of the datasets, and incomparably better results than CLIP. Also, we observe that on most attributes MUZE improves the performance over the baselines by a large margin, and even for the few attributes where it does not improve, the performance does not degrade by a significant amount.
\input{sections/main_tables}
\subsection{Behaviour Analysis}
\begin{figure}
    \centering
    {\includegraphics[clip, trim=15.6cm 6.9cm 15.5cm 7cm, width=0.50\textwidth]  {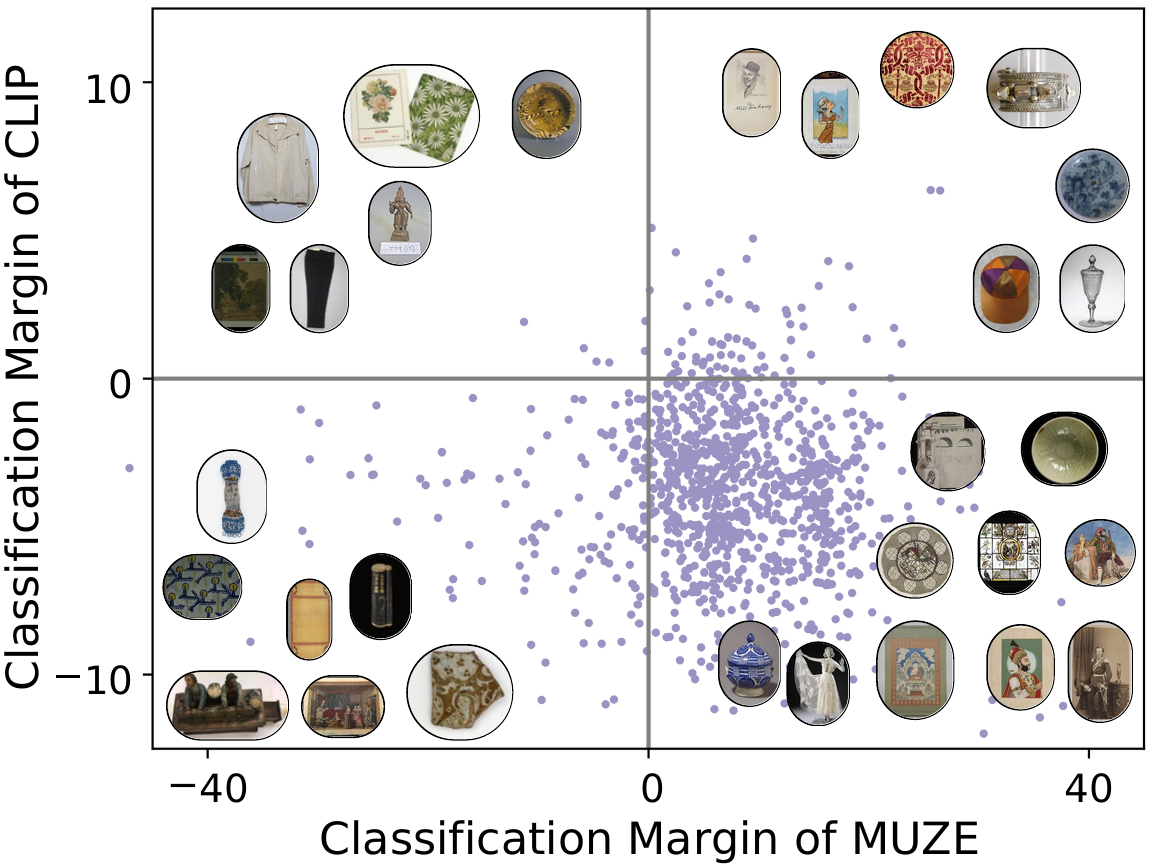}}
    {\includegraphics[clip, trim=0.2cm 0.3cm 0cm 0cm,width=0.48\textwidth]{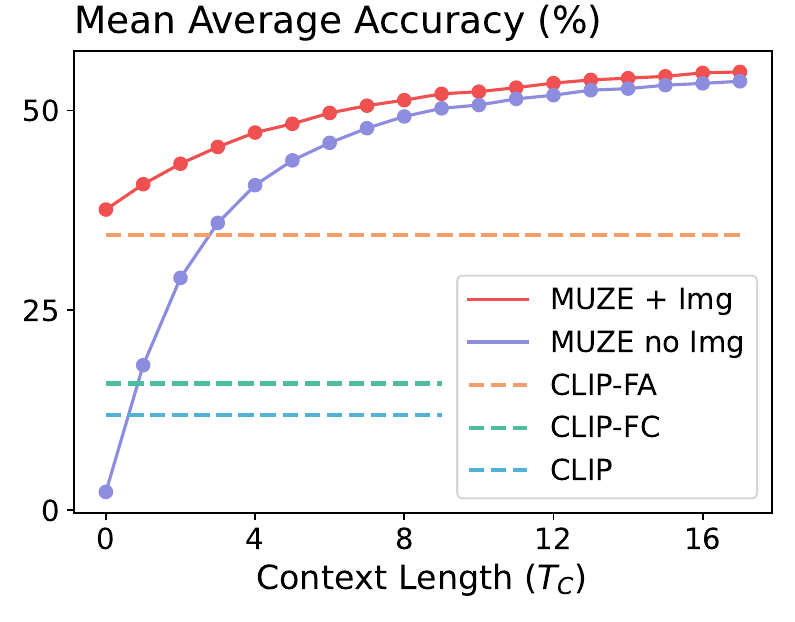}}
    \caption{Performance comparisons between MUZE and variants of CLIP; \textit{Left}: We compare the classification margin obtained by MUZE and CLIP over a batch of size 1024 from the validation set. We test both models on the \textit{categories} attribute and compute the margin as the score difference between the highest ranked correct and incorrect values. \textit{Right}: Comparison of scores obtained by MUZE for varying number of attribute-value pairs provided as input context ($\mathcal{T}_C$) when MUZE receives an image as input (red) vs when no image is provided (light purple). Also shown are the CLIP baselines which only process the image information ($\mathcal{T}_C=\emptyset$) }
    \label{fig:var_context}
    \vspace{-7mm}
\end{figure}

To analyse the easy and hard to predict examples, we compute the difference between highest score of a correct element and the highest score of an incorrect element for both CLIP and MUZE on predicting the values corresponding to the \textit{categories} attribute of A-MUZE. We plot a comparison of these scores in \cref{fig:var_context}, along with some images in each region for better analysis. We see that the distribution of the correct predicted values tend to move to the region where only MUZE was able to give the correct response.
We observed the hardest to guess images by both models are the ones depicting broken objects, or very small, made from glass, not very detailed. Also, there are very few objects which have better results on CLIP.

\subsection{Ablation Studies}
\subsubsection{MUZE base model variation}
Based on the different CLIP models (the original pre-trained model, and the ones obtained by fine-tuning) we train and evaluate different parseNet models, and we refer to them as MUZE-C, MUZE-CFC, MUZE-CFA, respectively. We present the results in \Cref{tab:MUZE_ablation}.
We observe that the best performance on A-MUZE is obtained by MUZE-CFC, while on B-MUZE the best variant is MUZE-CFA. However, the differences between the 3 variants are not very significant, showing that the training of parseNet is not heavily influenced by the CLIP variant used as backbone.

\begin{table}
\caption{Comparison of average results obtained by the three MUZE variants on the two data subsets. We report in each case the results averaged over all the attributes in the data subset.  }
\label{tab:MUZE_ablation}
\centering
\scriptsize
\begin{tabular}{lrrrrrr}
\toprule
Sub-Dataset   & \multicolumn{3}{c}{A-MUZE} & \multicolumn{3}{c}{B-MUZE} \\
\cmidrule(rl){2-4} \cmidrule(rl){5-7}  
Method        & {MUZE-C} & {MUZE-CFC}& {MUZE-CFA} & {MUZE-C} & {MUZE-CFC}& {MUZE-CFA} \\
\midrule
Mean Avg Prec     & 4.80           & \textbf{4.87}  & 4.36  & 2.47  & 2.43  & \textbf{2.98}  \\
Mean Avg Acc      & 23.47          & \textbf{23.51} & 22.70 & 29.78 & 29.87 & \textbf{30.36} \\
Mean Acc @ 1      & 22.67          & \textbf{22.81} & 21.94 & 29.57 & 29.67 & \textbf{30.47} \\
Mean Hit Rate @ 5 & \textbf{28.42} & 28.41          & 27.60 & 34.49 & 34.50 & \textbf{34.63} \\        
\bottomrule
\end{tabular}
\vspace{-2mm}
\end{table}

\vspace{-4mm}
\subsubsection{Impact of context length}
In \cref{fig:var_context} we show the variation of performance obtained when we varied the size of context given as input, $\mathcal{T}_c$. For this, we have trained two different models, one that receives the image along with the context, and the other which only gets the context. In order to be able to evaluate them with variable context length, we also trained them by randomly dropping some attribute-value pairs in the context. We compared these results with the results obtained by CLIP, CLIP-FC and CLIP-FA, which are all evaluated without context. For more details about attempts to use context with CLIP, see \cref{clip_context}. %

We observe the ascending trend of the performance as we increase the context, as well as the fact that having both image and context helps MUZE to achieve better performance than all other methods. We also see that in MUZE dataset, there are strong dependencies among the values of the different attributes.

%% file: sections/main_tables.tex
\begin{table}
\caption{Results on A-MUZE. We observe that results of MUZE-CFA (MUZE over CLIP-FA) are about 1.5 times better than the best CLIP method, and more than 3 times better than CLIP. The highest precision is obtained over classifiable attributes like \textit{categories}, \textit{objectType}, \textit{partTypes}, \textit{techniques}. }
\label{tab:vamus}
\centering
\scriptsize
\begin{tabular}{lrrrrrrrrrrrr}
\toprule
 \multicolumn{1}{c}{\multirow{3}{*}{Attribute}} & \multicolumn{4}{c}{Mean Avg. Prec} & \multicolumn{4}{c}{Mean Avg. Accuracy} & \multicolumn{4}{c}{Mean Acc @ 1} \\
\cmidrule(rl){2-5} \cmidrule(rl){6-9} \cmidrule(rl){10-13}
                           \multicolumn{1}{c}{}                           & {CLIP} & \shortstack{CLIP\\ FC}& \shortstack{CLIP\\ FA}& \shortstack{MUZE\\ CFA}& \multicolumn{1}{l}{CLIP} & \shortstack{CLIP\\ FC}& \shortstack{CLIP\\FA} & \shortstack{MUZE\\ CFA}& \multicolumn{1}{l}{CLIP} & \shortstack{CLIP\\ FC}& \shortstack{CLIP\\ FA}& \shortstack{MUZE\\ CFA}\\
\midrule
artistMakerPerson      & 0.33          & 0.33          & \textbf{1.63} & 1.01           & 0.13          & 0.10  & \textbf{6.44} & 0.25           & 0.01          & 0.01          & \textbf{5.69} & 0.01           \\
briefDescription       & 1.99          & 1.93          & 2.10          & \textbf{2.44}  & \textbf{3.42} & 2.88  & 3.22          & 1.08           & \textbf{5.05} & 3.51          & 3.83          & 0.81           \\
categories             & 5.71          & 6.09          & 8.34          & \textbf{12.27} & 11.96         & 15.84 & 34.46         & \textbf{58.43} & 8.71          & 12.75         & 40.13         & \textbf{74.95} \\
historicalContext      & 0.72          & \textbf{0.95} & 0.58          & 0.63           & 0.08          & 0.05  & 0.02          & \textbf{0.47}  & \textbf{0.02} & 0.01          & 0.01          & 0.01           \\
marksAndInscriptions   & \textbf{0.72} & 0.64          & 0.57          & 0.52           & 0.30          & 0.13  & 0.43          & \textbf{0.63}  & \textbf{0.21} & 0.08          & 0.01          & 0.02           \\
materials              & 1.42          & 1.56          & 4.39          & \textbf{6.01}  & 8.59          & 9.11  & 37.39         & \textbf{76.25} & 3.19          & 4.43          & 34.94         & \textbf{84.46} \\
materialsAndTechn. & 0.71          & 0.85          & 2.36          & \textbf{3.15}  & 0.82          & 0.74  & 2.67          & \textbf{7.44}  & 0.13          & 0.13          & 0.27          & \textbf{3.84}  \\
objectHistory          & 0.59          & 0.67          & \textbf{1.02} & 0.63           & 0.10          & 0.09  & \textbf{1.75} & 0.74           & 0.06          & 0.04          & \textbf{2.23} & 0.29           \\
objectType             & 3.93          & 4.42          & 7.77          & \textbf{17.21} & 2.16          & 1.43  & 1.40          & \textbf{6.32}  & 0.31          & 0.19          & 0.29          & \textbf{1.06}  \\
partTypes              & 3.08          & 3.35          & 4.95          & \textbf{13.57} & 0.58          & 0.33  & 0.29          & \textbf{0.85}  & 0.08          & 0.05          & 0.12          & \textbf{0.36}  \\
physicalDescription    & 1.76          & 1.88          & \textbf{2.17} & 2.06           & \textbf{4.40} & 3.22  & 2.55          & 2.57           & \textbf{3.89} & 2.27          & 1.45          & 0.73           \\
placesOfOrigin         & 1.97          & 1.38          & 5.00          & \textbf{9.64}  & 0.47          & 0.47  & 0.79          & \textbf{0.90}  & 0.03          & \textbf{0.03} & 0.01          & 0.01           \\
production             & \textbf{0.46} & 0.41          & 0.43          & 0.36           & 0.11          & 0.09  & 0.19          & \textbf{0.25}  & \textbf{0.06} & 0.05          & 0.00          & 0.00           \\
productionDates        & 0.44          & 0.49          & 2.98          & \textbf{3.41}  & 6.30          & 5.84  & 58.98         & \textbf{77.00} & 2.19          & 1.10          & 46.45         & \textbf{70.68} \\
styles                 & 1.27          & 0.85          & 3.45          & \textbf{4.68}  & 6.75          & 4.48  & 83.11         & \textbf{86.94} & 1.55          & 1.00          & 82.56         & \textbf{86.93} \\
summaryDescription     & 1.25          & \textbf{1.42} & 1.16          & 0.91           & 0.53          & 0.24  & 0.11          & \textbf{3.14}  & \textbf{0.77} & 0.16          & 0.02          & 0.02           \\
techniques             & 1.25          & 1.61          & 5.90          & \textbf{6.19}  & 6.44          & 6.30  & 44.47         & \textbf{75.17} & 4.07          & 4.62          & 36.09         & \textbf{80.00} \\
titles                 & 2.81          & 2.24          & 2.78          & \textbf{3.06}  & 3.09          & 1.89  & 2.47          & \textbf{24.72} & 2.17          & 1.11          & 1.34          & \textbf{6.38}  \\
                           \midrule
\textbf{Average}       & 1.69          & 1.73          & 3.20          & \textbf{4.87}  & 3.12          & 2.96  & 15.60         & \textbf{23.51} & 1.80          & 1.75          & 14.19         & \textbf{22.81} \\
\bottomrule
\end{tabular}
\end{table}

\begin{table}
\caption{Results on B-MUZE. We presented the results for MUZE-CFA compared to CLIP variants, and we note that MUZE has almost exclusively the best results, with great difference over the next method.}
\label{tab:britmus}
\centering
\scriptsize
\begin{tabular}{lrrrrrrrrrrrr}
\toprule
 \multicolumn{1}{c}{\multirow{2}{*}{Attribute}} & \multicolumn{4}{c}{Mean Avg Accuracy}                                                                                                                                                                                                               & \multicolumn{4}{c}{Mean Acc @ 1}                                                                                                                                                                                                                  & \multicolumn{4}{c}{Mean Hit Rate @ 5}                                                                                                                                                                                                             \\
\cmidrule(rl){2-5} \cmidrule(rl){6-9} \cmidrule(rl){10-13}
                           \multicolumn{1}{c}{}                           & \multicolumn{1}{l}{CLIP} & \shortstack{CLIP\\ FC}& \shortstack{CLIP\\ FA}& \shortstack{MUZE\\ CFA}& \multicolumn{1}{l}{CLIP} & \shortstack{CLIP\\ FC}& \shortstack{CLIP\\FA} & \shortstack{MUZE\\ CFA}& \multicolumn{1}{l}{CLIP} & \shortstack{CLIP\\ FC}& \shortstack{CLIP\\ FA}& \shortstack{MUZE\\ CFA}\\
\midrule
 Assoc name                                     & 0.08                     & 0.08                                                                  & 0.17                                                                  & \textbf{3.03}                                                          & 0.01                     & 0.02                                                                  & \textbf{0.04}                                                         & 0.02                                                                   & 0.05                     & 0.03                                                                  & \textbf{0.13}                                                         & 0.06                                                                   \\
                           Culture                                        & 3.47                     & 7.57                                                                  & 64.33                                                                 & \textbf{73.78}                                                         & 1.46                     & 5.18                                                                  & 63.86                                                                 & \textbf{81.37}                                                         & 5.27                     & 13.78                                                                 & 77.50                                                                 & \textbf{86.31}                                                         \\
                           Curators Comments                              & 0.12                     & 0.08                                                                  & 0.05                                                                  & \textbf{0.25}                                                          & \textbf{0.05}            & 0.02                                                                  & 0.01                                                                  & 0.00                                                                   & \textbf{0.24}            & 0.11                                                                  & 0.05                                                                  & 0.00                                                                   \\
                           Inscription                                    & 0.08                     & 0.08                                                                  & 0.27                                                                  & \textbf{0.67}                                                          & 0.02                     & 0.02                                                                  & \textbf{0.13}                                                         & 0.02                                                                   & 0.07                     & 0.09                                                                  & \textbf{0.25}                                                         & 0.06                                                                   \\
                           Materials                                      & 16.16                    & 18.94                                                                 & 76.83                                                                 & \textbf{79.38}                                                         & 8.71                     & 11.38                                                                 & 74.02                                                                 & \textbf{80.96}                                                         & 25.07                    & 28.26                                                                 & 88.30                                                                 & \textbf{88.42}                                                         \\
                           Object type                                    & 0.40                     & 0.27                                                                  & 0.29                                                                  & \textbf{0.83}                                                          & 0.08                     & 0.02                                                                  & 0.06                                                                  & \textbf{0.35}                                                          & 0.17                     & 0.11                                                                  & 0.16                                                                  & \textbf{0.88}                                                          \\
                           Producer name                                  & 0.09                     & 0.10                                                                  & \textbf{0.30}                                                         & 0.27                                                                   & 0.02                     & 0.02                                                                  & \textbf{0.03}                                                         & 0.01                                                                   & 0.06                     & 0.07                                                                  & \textbf{0.08}                                                         & 0.05                                                                   \\
                           Production date                                & 1.22                     & 0.88                                                                  & 42.60                                                                 & \textbf{48.74}                                                         & 0.40                     & 0.22                                                                  & 33.25                                                                 & \textbf{40.30}                                                         & 1.21                     & 0.69                                                                  & 50.29                                                                 & \textbf{56.72}                                                         \\
                           Production place                               & 0.48                     & 0.50                                                                  & 7.30                                                                  & \textbf{38.16}                                                         & 0.12                     & 0.07                                                                  & 3.14                                                                  & \textbf{32.82}                                                         & 0.34                     & 0.27                                                                  & 9.19                                                                  & \textbf{41.11}                                                         \\
                           Subjects                                       & 2.83                     & 2.94                                                                  & 24.82                                                                 & \textbf{52.68}                                                         & 1.58                     & 1.88                                                                  & 19.13                                                                 & \textbf{59.61}                                                         & 4.12                     & 5.15                                                                  & 39.33                                                                 & \textbf{64.33}                                                         \\
                           Technique                                      & 5.33                     & 6.81                                                                  & 54.56                                                                 & \textbf{66.22}                                                         & 3.05                     & 4.97                                                                  & 49.78                                                                 & \textbf{70.14}                                                         & 8.29                     & 11.09                                                                 & 72.31                                                                 & \textbf{77.60}                                                         \\
                           Title                                          & 0.13                     & 0.10                                                                  & \textbf{0.37}                                                         & 0.36                                                                   & \textbf{0.06}            & 0.02                                                                  & 0.05                                                                  & 0.01                                                                   & \textbf{0.11}            & 0.06                                                                  & 0.08                                                                  & 0.02                                                                   \\
                           \midrule
                           \textbf{Average}                               & 2.53                     & 3.20                                                                  & 22.66                                                                 & \textbf{30.36}                                                         & 1.30                     & 1.98                                                                  & 20.29                                                                 & \textbf{30.47}                                                         & 3.75                     & 4.98                                                                  & 28.14                                                                 & \textbf{34.63} \\                                               
                    \bottomrule
\end{tabular}
\end{table}

%% file: sections/conclusion.tex
\vspace{-2mm}
\section{Conclusion}
\vspace{-2mm}
We studied the problem of fine-grained and structured understanding of the museum exhibits using the pre-trained embedding of the CLIP model. 
For the structured understanding, we proposed a method that learns to provide the answers to the query attributes, from the input image, thus generating the final output in a tabular format. When a part of the tabular information is already known, we enable our method to answer better by leveraging the available additional context. To develop such methods, we also gathered, curated, and benchmarked a dataset of 200K+ image-table pairs, which we will make publicly available. Our proposed method is shown to be effective using exhaustive experiments and analysis. Our work fosters further research and application development in fine-grained understanding museum exhibits in a structured manner, and develop methods that can be used potentially beyond the museum applications. 
\vspace{-2mm}\subsubsection{Acknowledgements}This research was partially funded by the Ministry of Education and Science of Bulgaria (support for INSAIT, part of the Bulgarian National Roadmap for Research Infrastructure). We thank the Bulgarian National Archaeological Institute with Museum for the support and guidance, the British Museum and Victoria\&Albert Museum for the access to their data that made this research possible, and Google DeepMind which provided vital support and resources for this research. We also thank the anonymous reviewers for their efforts and valuable feedback to improve our work.

%% file: sections/appendix.tex
\clearpage
\appendix
\section{Appendix Overview}
In \Cref{distrib} we further analyse the properties of different attributes in the two data subsets.
In \Cref{quality_checks} we present more details about the quality of the labels associated with each image.
In \Cref{nomenclature} we offer more explanation about our notations.
In \Cref{clip_context} we present the results obtained by using CLIP with textual information.
In \Cref{finetuning_text} we present the results obtained by finetuning CLIP's text encoder.
In \Cref{finetuning_phrase} we present the results obtained by finetuning CLIP using a phrase-like input.
In \Cref{goodbad1} we compare the scores of MUZE and CLIP-FA. 
In \Cref{subsec:muze_var} we show the results of MUZE ablations over all attributes from both data subsets.
In \Cref{goodbad2} we provide multiple image examples of attribute prediction for MUZE, CLIP and CLIP-FA.

\vspace{-0.2cm}
\section{Dataset attribute distribution}
\label{distrib}
\begin{figure}
\vspace{-0.3cm}
    \centering
    {\includegraphics[clip, trim=0cm 0cm 0cm 0cm, width=0.8\textwidth]  {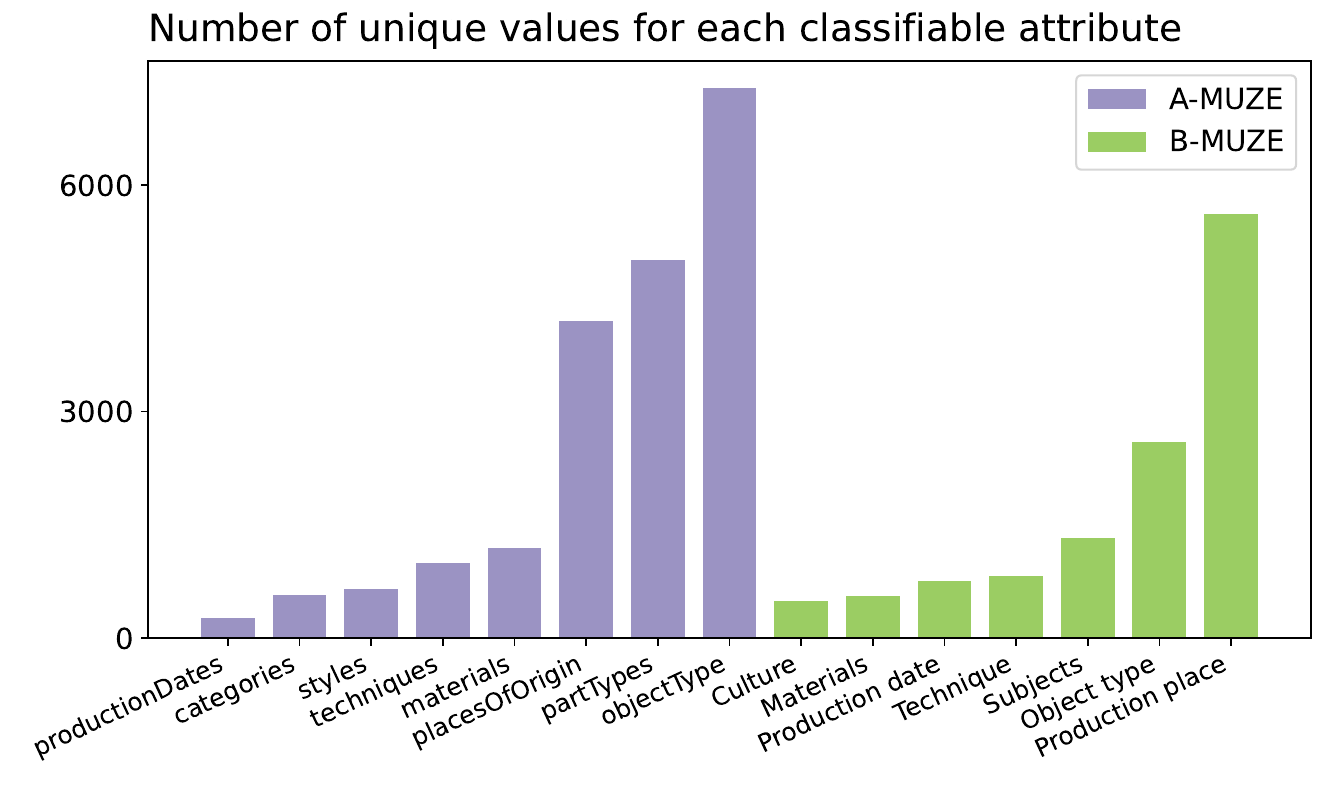}}
    \caption{Comparison between the number of unique values for each classifiable attribute from MUZE dataset in each data subset.  }
    \label{fig:unique_values}
\end{figure}

For a better understanding of the dataset, we checked the composition of each attribute from both data subsets. We analysed the amount of unique values from classifiable attributes, and we observed attributes like \textit{categories, materials, Subjects, Techniques} tend to have few unique values, while there are also classifiable attributes with thousands of different values, see \cref{fig:unique_values}.

\vspace{-0.5cm}
\section{Experts' data quality checks \& release}
\label{quality_checks}

For the collection of image/text pairs, we took data from Victoria and Albert Museum (A-MUZE) and British Museum (B-MUZE).

We took 500 image-text pairs from the MUZE dataset, and asked 3 experts specialized in Museum Art \& History to rate how well the image matched the text on a scale from 1 to 5, (1 = they totally disagreed, 5 = they totally agreed).
The average score from all three experts was \textbf{4.507}, showing strong agreements. 

\begin{figure}
    \centering
    \vspace{-7mm}
    {\includegraphics[clip, trim=0.7cm 1.0cm 2.1cm 1.1cm, width=.5\linewidth]    {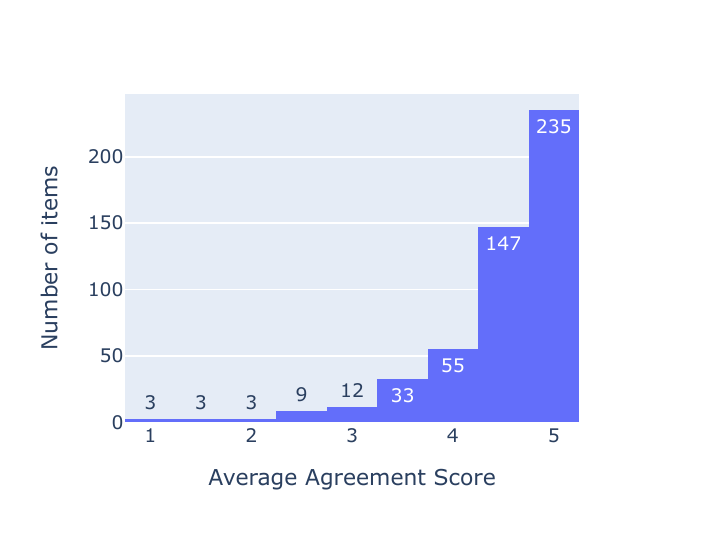}}
    \caption{Experts Label agreement.}
    \label{fig:experts_score}
    \vspace{-3.5mm}
\end{figure}

See the experts' rating distribution in \cref{fig:experts_score}. The experts considered 9 exhibits/images not to be in good agreement with the text. These exhibits were rated so because the objects were hard to photograph due to their material, were hard to recognize, or they came with a controversial historical background, where subjectivity may influence someone's opinion.

Following LAION-400M, we plan to release our curation under the CCBY-NC-4.0, which will include a direct web-link to the images. The images will remain under their copyright. Please, note that most museums (e.g. British) make their images available under the CCBY-NC-4.0 as well.

\vspace{-8mm}
\section{Nomenclature}
\label{nomenclature}
\vspace{-1mm}
To ensure clarity, consistency in terminology and an easier following of the MUZE-C/CFC/CFA notations meaning we provides definitions and explanations for key terms in the following nomenclature \cref{tab:nomenclature}.

\begin{table}[h!]
\caption{Notations explanation }
\label{tab:nomenclature}
\centering
\scriptsize
\begin{tabular}{cc}
Notation & Explanation                                          \\ \hline
CLIP     & frozen image encoder \& frozen text encoder          \\ 
CLIP-FC  & CLIP finetuned with captions created from attributes \\
CLIP-FA  & CLIP finetuned with each attribute value             \\
MUZE-C   & MUZE built over CLIP backbone                        \\
MUZE-CFC & MUZE built over CLIP-FC backbone                     \\
MUZE-CFA & MUZE built over CLIP-FA backbone                    
\end{tabular}
\end{table}

\vspace{-5mm}
\section{CLIP with context}
\label{clip_context}
In some of CLIP testing variation, we give context to CLIP during inference. 
When we do inference on CLIP without context we match the image embedding with the embedding of each value for the attributes.
For CLIP with context, called CLIP-CTX, we build the context by concatenating the known attribute values of each instance as a suffix, separated by comma, for each known attribute except for the query attribute. For the inference for CLIP-CTX, we created all possible concatenations between the context and each possible value of the query attribute and compute the similarity with the image embedding. We present the results in \Cref{tab:clip_context_amuze,tab:clip_context_bmuze}.
We observe the results of CLIP-CTX are not as good as the other CLIP variants, having big difference to CLIP-FA.

\begin{table}[]
\vspace{-2mm}
\caption{Comparison of the results obtained by giving context information to CLIP and the other variants of CLIP, including CLIP fine-tuned. We report in each case the results over classifiable attributes of A-MUZE data subset.  }
\label{tab:clip_context_amuze}
\scriptsize
\centering
\begin{tabular}{lrrrrrrrrrrrr}
\toprule
\multicolumn{1}{c}{\multirow{2}{*}{Attribute}} & \multicolumn{4}{c}{\textbf{Mean Avg Prec}}                                                                                                                                                                                                        & \multicolumn{4}{c}{\textbf{Mean Avg Recall}}                                                                                                                                                                                                      & \multicolumn{4}{c}{\textbf{Mean Acc @ 1}}                                                                                                                                                                                                         \\
\cmidrule(rl){2-5} \cmidrule(rl){6-9} \cmidrule(rl){10-13}
\multicolumn{1}{c}{}                           & \multicolumn{1}{l}{CLIP} & \multicolumn{1}{l}{\begin{tabular}[c]{@{}l@{}}CLIP\\ FC\end{tabular}} & \multicolumn{1}{l}{\begin{tabular}[c]{@{}l@{}}CLIP\\ FA\end{tabular}} & \multicolumn{1}{l}{\begin{tabular}[c]{@{}l@{}}CLIP\\ CTX\end{tabular}} & \multicolumn{1}{l}{CLIP} & \multicolumn{1}{l}{\begin{tabular}[c]{@{}l@{}}CLIP\\ FC\end{tabular}} & \multicolumn{1}{l}{\begin{tabular}[c]{@{}l@{}}CLIP\\ FA\end{tabular}} & \multicolumn{1}{l}{\begin{tabular}[c]{@{}l@{}}CLIP\\ CTX\end{tabular}} & \multicolumn{1}{l}{CLIP} & \multicolumn{1}{l}{\begin{tabular}[c]{@{}l@{}}CLIP\\ FC\end{tabular}} & \multicolumn{1}{l}{\begin{tabular}[c]{@{}l@{}}CLIP\\ FA\end{tabular}} & \multicolumn{1}{l}{\begin{tabular}[c]{@{}l@{}}CLIP\\ CTX\end{tabular}} \\
\midrule
categories                                     & 5.71                     & 6.09                                                                  & \textbf{8.34}                                                         & 1.745                                                                  & 11.96                    & 15.84                                                                 & \textbf{34.46}                                                        & 3.40                                                                   & 8.71                     & 12.75                                                                 & \textbf{40.13}                                                        & 0.54                                                                   \\
materials                                      & 1.42                     & 1.56                                                                  & \textbf{4.39}                                                         & 0.250                                                                  & 8.59                     & 9.11                                                                  & \textbf{37.39}                                                        & 3.50                                                                   & 3.19                     & 4.43                                                                  & \textbf{34.94}                                                        & 0.40                                                                   \\
objectType                                     & 3.93                     & 4.42                                                                  & \textbf{7.77}                                                         & 0.041                                                                  & 2.16                     & 1.43                                                                  & \textbf{1.40}                                                         & 0.02                                                                   & 0.31                     & 0.19                                                                  & \textbf{0.29}                                                         & 0.00                                                                   \\
productionDates                                & 0.44                     & 0.49                                                                  & \textbf{2.98}                                                         & 1.128                                                                  & 6.30                     & 5.84                                                                  & \textbf{58.98}                                                        & 4.14                                                                   & 2.19                     & 1.10                                                                  & \textbf{46.45}                                                        & 2.36                                                                   \\
styles                                         & 1.27                     & 0.85                                                                  & \textbf{3.45}                                                                 & 0.153                                                                  & 6.75                     & 4.48                                                                  & \textbf{83.11}                                                                 & 4.34                                                                   & 1.55                     & 1.00                                                                  & \textbf{82.56}                                                                 & 0.56                                                                   \\
techniques                                     & 1.25                     & 1.61                                                                  & \textbf{5.90}                                                                  & 0.501                                                                  & 6.44                     & 6.30                                                                  & \textbf{44.47}                                                                 & 1.68                                                                   & 4.07                     & 4.62                                                                  & \textbf{36.09}                                                                 & 0.42                                    \\ 
\bottomrule
\end{tabular}
\vspace{-1.5mm}
\end{table}

\vspace{-3.5mm}

\begin{table}[]
\vspace{-5mm}
\caption{Comparison between giving context information to CLIP
and the other variants of CLIP. We report in each case the
results over classifiable attributes of B-MUZE data subset.  }
\label{tab:clip_context_bmuze}
\scriptsize
\centering
\begin{tabular}{lrrrrrrrrrrrr}
\toprule

\multicolumn{1}{c}{\multirow{2}{*}{Attribute}} & \multicolumn{4}{c}{\textbf{Mean Avg Prec}}                                                                                                                                                                                                            & \multicolumn{4}{c}{\textbf{Mean Avg Recall}}                                                                                                                                                                                                          & \multicolumn{4}{c}{\textbf{Mean Acc @ 1}}                                                                                                                                                                                                             \\
\cmidrule(rl){2-5} \cmidrule(rl){6-9} \cmidrule(rl){10-13}
\multicolumn{1}{c}{}                           & \multicolumn{1}{l}{CLIP} & \multicolumn{1}{l}{\begin{tabular}[c]{@{}l@{}}CLIP\\ FC\end{tabular}} & \multicolumn{1}{l}{\begin{tabular}[c]{@{}l@{}}CLIP\\ FC\end{tabular}} & \multicolumn{1}{l}{\begin{tabular}[c]{@{}l@{}}CLIP\\ CTX\end{tabular}} & \multicolumn{1}{l}{CLIP} & \multicolumn{1}{l}{\begin{tabular}[c]{@{}l@{}}CLIP\\ FC\end{tabular}} & \multicolumn{1}{l}{\begin{tabular}[c]{@{}l@{}}CLIP\\ FC\end{tabular}} & \multicolumn{1}{l}{\begin{tabular}[c]{@{}l@{}}CLIP\\ CTX\end{tabular}} & \multicolumn{1}{l}{CLIP} & \multicolumn{1}{l}{\begin{tabular}[c]{@{}l@{}}CLIP\\ FC\end{tabular}} & \multicolumn{1}{l}{\begin{tabular}[c]{@{}l@{}}CLIP\\ FC\end{tabular}} & \multicolumn{1}{l}{\begin{tabular}[c]{@{}l@{}}CLIP\\ CTX\end{tabular}} \\
\midrule
Culture                                        & 0.98                     & 1.32                                                                  & \textbf{6.55}                                                         & 0.44                                                                       & 3.47                     & 7.57                                                                  & \textbf{64.33}                                                        & 2.78                                                                       & 1.46                     & 5.18                                                                  & \textbf{63.86}                                                        & 1.03                                                                       \\
Materials                                      & 1.55                     & 1.70                                                                  & \textbf{8.31}                                                         & 0.29                                                                       & 16.16                    & 18.94                                                                 & \textbf{76.83}                                                        & 4.22                                                                       & 8.71                     & 11.38                                                                 & \textbf{74.02}                                                        & 1.66                                                                       \\
Subjects                                       & 1.81                     & 0.79                                                                  & \textbf{3.35}                                                         & 0.18                                                                       & 2.83                     & 2.94                                                                  & \textbf{24.82}                                                        & 0.63                                                                       & 1.58                     & 1.88                                                                  & \textbf{19.13}                                                        & 0.94                                                                       \\
Technique                                      & 1.50                     & 1.31                                                                  & \textbf{5.69}                                                         & 0.27                                                                       & 5.33                     & 6.81                                                                  & \textbf{54.56}                                                        & 1.24                                                                       & 3.05                     & 4.97                                                                  & \textbf{49.78}                                                        & 0.06                         \\
\bottomrule
\end{tabular}
\vspace{-1.5mm}
\end{table}

\vspace{-3.5mm}
\section{Finetuning text encoder}
\label{finetuning_text}
We conducted an experiment by finetuning the CLIP text encoder (CLIP-Ftext), which is reported in \cref{tab:MUZE_other_ablations}. We observed that finetuning the text encoder deteriorates the results from CLIP-FA.

\begin{table}[]
\caption{Average results of CLIP experiments and MUZE. }
\label{tab:MUZE_other_ablations}
\centering
\scriptsize
\begin{tabular}{c|c|c|c|c|c}
CLIP&CLIP-phr(inf)&CLIP-FA&CLIP-Fphrase&CLIP-Ftext&MUZE-CFA\\ \hline
1.69 & 1.22  & 3.20    & 2.61         & 2.45       & 4.87    
\end{tabular}
\vspace{-4mm}
\end{table}

\section{Finetuning CLIP using phrase-like input}
\label{finetuning_phrase}We conducted an experiment using the phrase format ``the \{key\} of the object are'' as input for CLIP, which we denoted it CLIP-Fphrase. We reported the results in \cref{tab:MUZE_other_ablations}, and we noted that the input format does not improve the results of CLIP-FA. We believe more fine-grained and exhibit specific phrases are needed for phrase-like input format to be useful in our settings, which are not trivial at the moment.

\vspace{-.2cm}
\section{Comparison between MUZE and CLIP-FA over good/bad results}
\label{goodbad1}
\begin{figure}[!h]
\vspace{-5mm}
    \centering
    {\includegraphics[clip, trim=0cm 0cm 0cm 0cm, width=0.38\textwidth]  {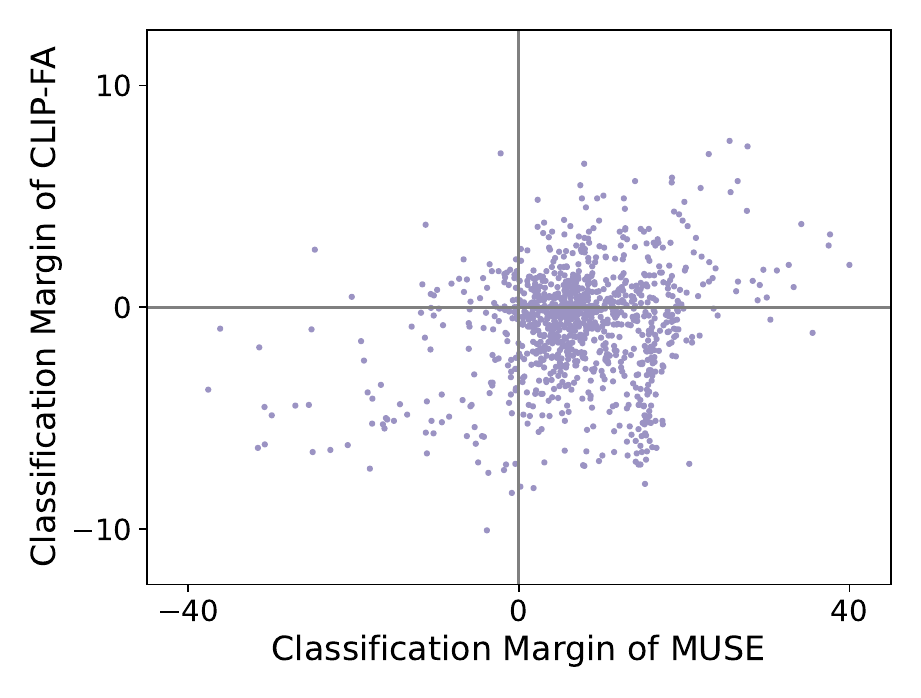}}
    {\includegraphics[clip, trim=0cm 0cm 0cm 0cm, width=0.38\textwidth]  {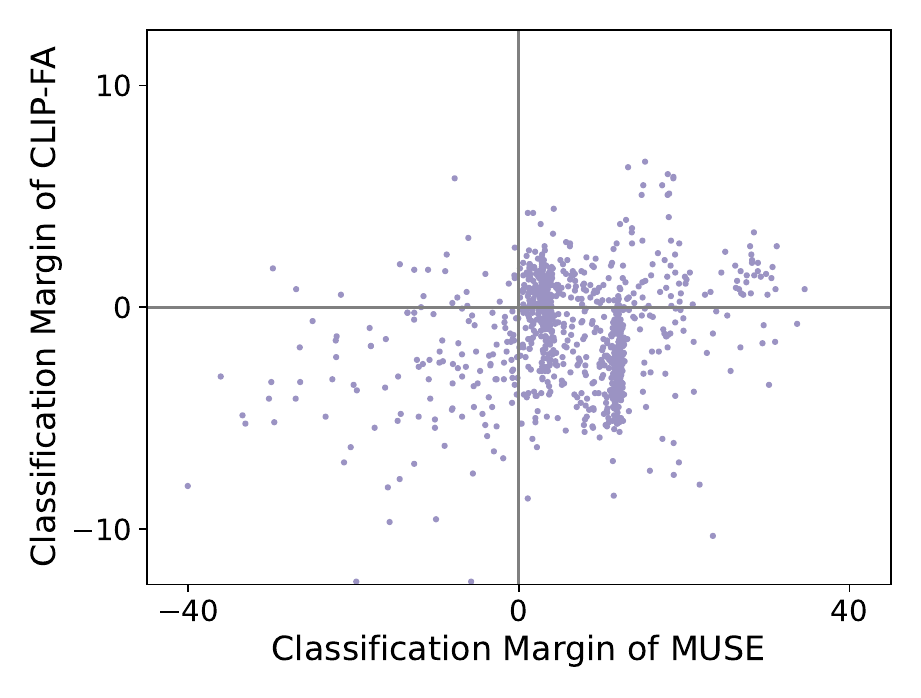}} \\
    {\includegraphics[clip, trim=0cm 0cm 0cm 0cm, width=0.38\textwidth]  {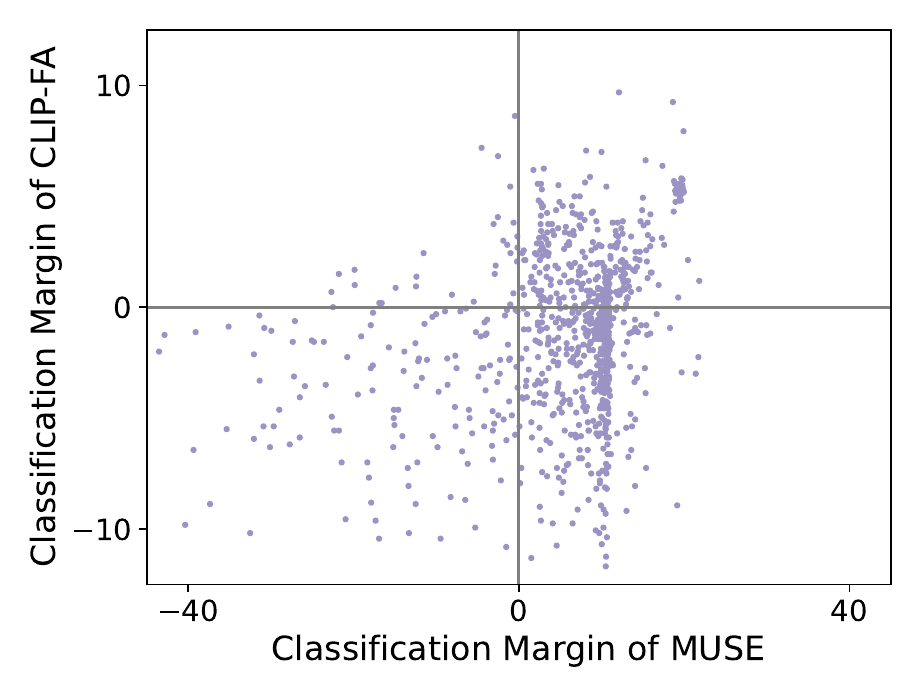}}
    {\includegraphics[clip, trim=0cm 0cm 0cm 0cm, width=0.38\textwidth]  {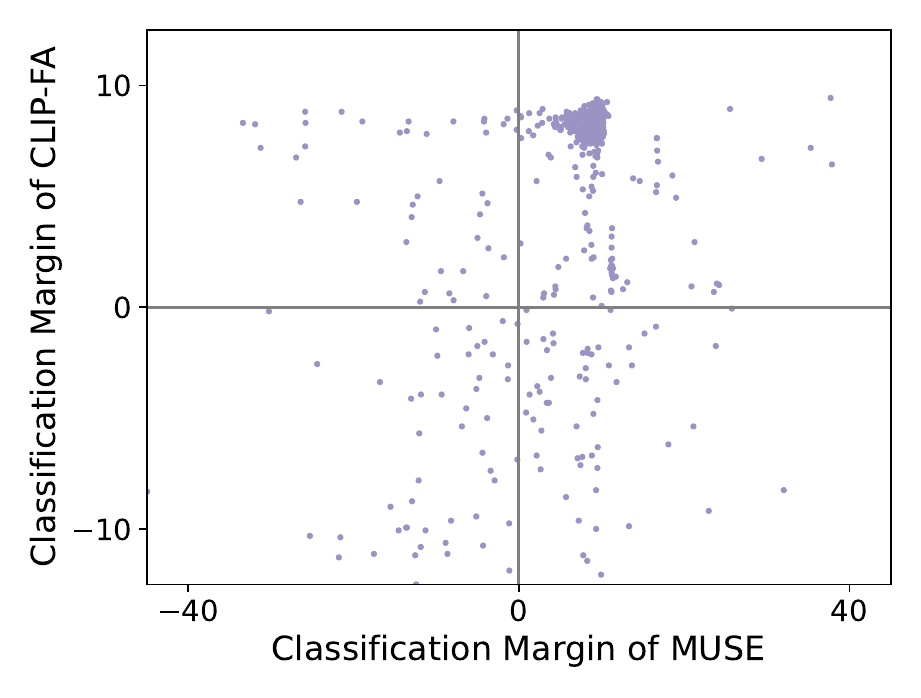}}
    \caption{We compare the classification margin obtained by MUZE and CLIP-FA over 1024 samples
from the validation set. We test both models on the \textit{categories}, \textit{materials} (up), \textit{technologies} and \textit{styles} (down) attributes and compute
the margin as the score difference between the highest ranked correct
and incorrect values.
}
    \label{fig:margins}
\vspace{-4mm}
\end{figure}

We provide here more examples similar to \cref{fig:var_context} (left) from the paper. We compared this time the classification margins of MUZE and CLIP-FA  , see \cref{fig:margins}.

\section{MUZE variants over A-MUZE and B-MUZE}
\label{subsec:muze_var}

In \Cref{tab:vamus_MUZE,tab:muze_variants2} we present the detailed results of the three MUZE variants on the two data subsets, over all the attributes. We notice that, for A-MUZE we have very good results by both MUZE-CFA and MUZE-CFC with small difference between the two methods, with the majority having slightly better results for MUZE-CFA. In the case of B-MUZE, the best results are all the time obtained using MUZE-CFA method.

\begin{table}[]
\caption{Comparison of average results obtained by the three MUZE variants on A-MUZE data subset. We report the results over all the attributes in the data subset.  }
\label{tab:vamus_MUZE}
\centering
\scriptsize
\begin{tabular}{lrrrrrrrrr}
\toprule
 \multicolumn{1}{c}{\multirow{3}{*}{Attribute}} & \multicolumn{3}{c}{Mean Avg. Prec} & \multicolumn{3}{c}{Mean Avg. Accuracy} & \multicolumn{3}{c}{Mean Acc @ 1}  \\
\cmidrule(rl){2-4} \cmidrule(rl){5-7} \cmidrule(rl){8-10} 
 & \makecell{MUZE \\ CLIP} & \makecell{MUZE\\ CFC} & \makecell{MUZE\\ CFA} 
 & \makecell{MUZE \\ CLIP} & \makecell{MUZE\\ CFC}& \makecell{MUZE\\ CFA}
 & \makecell{MUZE \\ CLIP}& \makecell{MUZE\\ CFC}& \makecell{MUZE\\ CFA} \\
\midrule
artistMakerPerson      & 1.05           & 1.01           & \textbf{1.18} & 0.26           & 0.25           & \textbf{0.29}  & 0.00           & 0.01           & \textbf{0.04}  \\
briefDescription       & 2.46           & 2.44           & \textbf{2.50} & 0.98           & 1.08           & \textbf{1.08}  & 0.65           & \textbf{0.81}  & 0.78           \\
categories             & 11.82          & \textbf{12.27} & 12.17         & 57.67          & 58.43          & \textbf{58.75} & 73.44          & 74.95          & \textbf{75.15} \\
historicalContext      & 0.48           & \textbf{0.63}  & 0.57          & 0.47           & 0.47           & 0.47           & 0.00           & \textbf{0.01}  & 0.00           \\
marksAndInscriptions   & 0.55           & 0.52           & \textbf{0.75} & 0.61           & \textbf{0.63}  & 0.62           & 0.00           & \textbf{0.02}  & \textbf{0.02}  \\
materials              & \textbf{6.17}  & 6.01           & 5.95          & \textbf{76.36} & 76.25          & 76.28          & \textbf{84.60} & 84.46          & 84.36          \\
materialsAndTechniques & \textbf{3.16}  & 3.15           & 2.62          & 7.41           & \textbf{7.44}  & 7.17           & 3.87           & 3.84           & \textbf{3.96}  \\
objectHistory          & 0.59           & 0.63           & \textbf{0.75} & \textbf{0.78}  & 0.74           & 0.55           & \textbf{0.33}  & 0.29           & 0.11           \\
objectType             & \textbf{17.38} & 17.21          & 13.28         & \textbf{6.50}  & 6.32           & 5.72           & \textbf{1.15}  & 1.06           & 0.97           \\
partTypes              & 13.43          & \textbf{13.57} & 12.34         & 0.82           & 0.85           & \textbf{0.89}  & 0.34           & \textbf{0.36}  & 0.36           \\
physicalDescription    & 2.06           & 2.06           & \textbf{2.12} & 2.49           & 2.57           & \textbf{2.62}  & 0.63           & 0.73           & \textbf{0.73}  \\
placesOfOrigin         & 8.72           & \textbf{9.64}  & 6.16          & 0.90           & \textbf{0.90}  & 0.85           & \textbf{0.01}  & 0.01           & 0.01           \\
production             & 0.27           & 0.36           & \textbf{0.42} & 0.25           & 0.25           & 0.25           & 0.00           & 0.00           & 0.00           \\
productionDates        & 2.89           & \textbf{3.41}  & 2.43          & 76.92          & \textbf{77.00} & 73.47          & 70.06          & \textbf{70.68} & 66.24          \\
styles                 & 4.96           & 4.68           & \textbf{6.17} & 86.67          & \textbf{86.94} & 85.01          & 86.68          & \textbf{86.93} & 84.90          \\
summaryDescription     & 0.85           & 0.91           & \textbf{0.96} & 3.14           & 3.14           & \textbf{3.15}  & 0.01           & \textbf{0.02}  & 0.01           \\
techniques             & \textbf{6.65}  & 6.19           & 4.94          & \textbf{75.53} & 75.17          & 66.46          & \textbf{80.06} & 80.00          & 70.71          \\
titles                 & 3.00           & 3.06           & \textbf{3.19} & 24.63          & 24.72          & \textbf{24.91} & 6.24           & 6.38           & \textbf{6.52}  \\
                           \midrule
\textbf{Average}       & 4.80           & \textbf{4.87}  & 4.36          & 23.47          & \textbf{23.51} & 22.70          & 22.67          & \textbf{22.81} & 21.94         \\
\bottomrule
\end{tabular}
\end{table}

\begin{table}[]
\caption{Comparison of average results obtained by the three MUZE variants on B-MUZE data subset. We report the results over all the attributes in the data subset.   }
\label{tab:muze_variants2}
\scriptsize
\centering
\begin{tabular}{lrrrrrrrrr}
\toprule
\multicolumn{1}{c}{\multirow{2}{*}{Attribute}} & \multicolumn{3}{c}{Mean Avg Prec}                                                                                                                                                                                         & \multicolumn{3}{c}{Mean Avg Recall}                                                                                                                                                                                       & \multicolumn{3}{c}{Mean Acc @ 1}                                                                                                                                                                                          \\
\cmidrule(rl){2-4} \cmidrule(rl){5-7} \cmidrule(rl){8-10}
\multicolumn{1}{c}{}                           & \multicolumn{1}{l}{\begin{tabular}[c]{@{}l@{}}MUSE\\ CLIP\end{tabular}} & \multicolumn{1}{l}{\begin{tabular}[c]{@{}l@{}}MUSE\\ CFC\end{tabular}} & \multicolumn{1}{l}{\begin{tabular}[c]{@{}l@{}}MUSE\\ CFA\end{tabular}} & \multicolumn{1}{l}{\begin{tabular}[c]{@{}l@{}}MUSE\\ CLIP\end{tabular}} & \multicolumn{1}{l}{\begin{tabular}[c]{@{}l@{}}MUSE\\ CFC\end{tabular}} & \multicolumn{1}{l}{\begin{tabular}[c]{@{}l@{}}MUSE\\ CFA\end{tabular}} & \multicolumn{1}{l}{\begin{tabular}[c]{@{}l@{}}MUSE\\ CLIP\end{tabular}} & \multicolumn{1}{l}{\begin{tabular}[c]{@{}l@{}}MUSE\\ CFC\end{tabular}} & \multicolumn{1}{l}{\begin{tabular}[c]{@{}l@{}}MUSE\\ CFA\end{tabular}} \\
\midrule
Assoc name        & 1.02          & 1.02 & \textbf{1.24} & 2.94          & 2.85          & \textbf{3.03}  & 0.02           & 0.02           & \textbf{0.02}  \\
Culture           & 6.12          & 7.30 & \textbf{7.49} & 73.51         & 73.68         & \textbf{73.78} & 81.48          & \textbf{81.73} & 81.37 \\
Curators Comments & 0.16 & 0.16 & \textbf{0.18} & 0.23 & 0.24          & \textbf{0.25}  & 0.00  & 0.00           & 0.00           \\
Inscription       & 0.38          & 0.43 & \textbf{0.45} & 0.55 & 0.55 & \textbf{0.67}  & 0.01           & 0.01           & \textbf{0.02}  \\
Materials         & 5.62          & 5.24 & \textbf{6.84} & 77.45         & 78.16         & \textbf{79.38} & 78.47          & 79.68          & \textbf{80.96} \\
Object type       & 3.46          & 2.97 & \textbf{5.74} & 0.75          & 0.77 & \textbf{0.83}  & 0.36           & \textbf{0.39}  & 0.35           \\
Producer name     & 1.36          & 1.42 & \textbf{1.66} & 0.27          & 0.27          & \textbf{0.27}  & 0.01           & 0.00           & \textbf{0.01}  \\
Production date   & 2.56          & 2.41 & \textbf{2.81} & 45.18         & 45.97         & \textbf{48.74} & 32.34          & 33.53          & \textbf{40.30} \\
Production place  & 2.38          & 2.25 & \textbf{2.76} & 38.02         & 37.96         & \textbf{38.16} & \textbf{33.01} & 32.72          & 32.82 \\
Subjects          & 1.87 & 1.61 & \textbf{1.90} & 52.38         & 51.76         & \textbf{52.68} & 59.35 & 58.17          & \textbf{59.61} \\
Technique         & 3.40          & 3.36 & \textbf{3.74} & 65.71         & 65.83         & \textbf{66.22} & 69.77          & 69.74          & \textbf{70.14} \\
Title             & \textbf{1.34} & 0.97 & 0.90          & \textbf{0.38} & 0.36          & 0.36           & \textbf{0.02}  & 0.01           & 0.01           \\
\midrule
\textbf{Average}                               & 2.47                                                                    & 2.43                                                                   & \textbf{2.98}                                                          & 29.78                                                                   & 29.87                                                                  & \textbf{30.36}                                                         & 29.57                                                                   & 29.67                                                                  & \textbf{30.47}     \\                \bottomrule                         
\end{tabular}
\end{table}

\section{Images examples: GOOD and BAD results}
\label{goodbad2}
In this section we present an analysis of the images for which MUZE algorithm manages to obtain results that are very close to the ground truth (GT), displayed with Prediction Type = GOOD, or not so close to the GT, displayed with Prediction Type = BAD. We compare the result of MUZE-CFC (MUZE) with CLIP and CLIP-FA and we analyse multiple classifiable attributes from A-MUZE data subset. For each sample we show all ground truth values and top two predictions of each method. We considered the prediction as GOOD if the first prediction of MUZE was found in the list of ground truth values.
Examples for more attributes can be found inside \texttt{image\_examples/<attribute\_name>}. Also, examples of images' descriptions can be found at \texttt{image\_examples/descriptions}.

\begin{table}[]
\caption{Results of MUZE and CLIP variants compared with ground truth (GT) over the attribute: \textit{categories} }
\label{tab:images1}
\scriptsize
\centering
\begin{tabular}{c|c|c|c|c|c|c|}
{Image} & Attribute                      & GT                                & MUZE                      & CLIP                      & CLIP-FA &Pred. Type \\
\midrule
\raisebox{-0.5\totalheight}{\includegraphics[clip, trim=0cm 1cm 0cm 1cm, width=0.15\textwidth]  {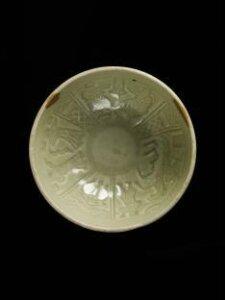}}
                
& {categories} 
& {\makecell{ceramics\\ stoneware}} 
& {\makecell{ceramics\\ stoneware}} 
& {\makecell{pottery\\ stoneware}}
& {\makecell{pottery\\ slipware}}   
& GOOD
\\
\raisebox{-0.5\totalheight}{\includegraphics[clip, trim=0cm 1cm 0cm 1cm, width=0.15\textwidth]  {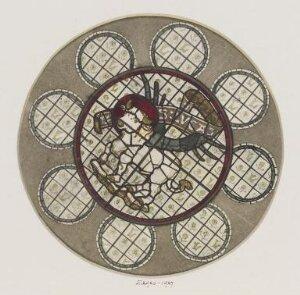}}
                
& {categories} 
& {\makecell{artcraft\\ stained\\ drawings\\ glass}} 
& {\makecell{ glass\\ stained}} 
& {\makecell{2003\\ embroideries}}
& {\makecell{glass\\ embroidery}}   
& GOOD

\\

\raisebox{-0.5\totalheight}{\includegraphics[clip, trim=0cm 1cm 0cm 1cm, width=0.15\textwidth]  {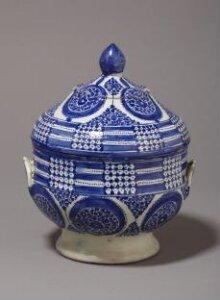}}
                
& {categories} 
& {\makecell{africa\\ ceramics}} 
& {\makecell{ ceramics\\ earthenware}} 
& {\makecell{pearlware\\ delftware}}
& {\makecell{pottery\\ slipware}}   
& GOOD
\\
\raisebox{-0.5\totalheight}{\includegraphics[clip, trim=0cm 1cm 0cm 1cm, width=0.15\textwidth]  {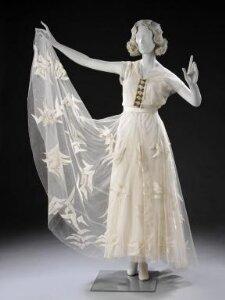}}
               
& {categories} 
& {\makecell{fashion}} 
& {\makecell{fashion\\ wear}} 
& {\makecell{fancy-dress\\ couture}}
& {\makecell{womenswear\\ couture}}   
& GOOD
\\
\raisebox{-0.5\totalheight}{\includegraphics[clip, trim=0cm 1cm 0cm 1cm, width=0.15\textwidth]  {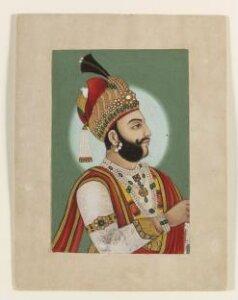}}
                
& {categories} 
& {\makecell{portraits\\ trust\\ royalty\\ indian\\ paintings}} 
& {\makecell{paintings\\ indian}} 
& {\makecell{persia\\ indian}}
& {\makecell{indian\\ paintings}}   
& GOOD
\\
\midrule
\raisebox{-0.5\totalheight}{\includegraphics[clip, trim=0cm 1cm 0cm 1cm, width=0.15\textwidth]  {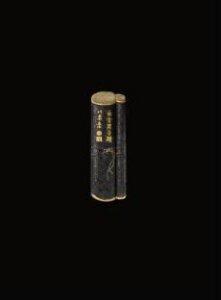}}
                
& {categories} 
& {\makecell{stationery}} 
& {\makecell{unknown\\ accessories}} 
& {\makecell{china\\ chinese}}
& {\makecell{lacquerware\\ instruments}}   
& BAD
\\
\raisebox{-0.5\totalheight}{\includegraphics[clip, trim=0cm 1cm 0cm 1cm, width=0.15\textwidth]  {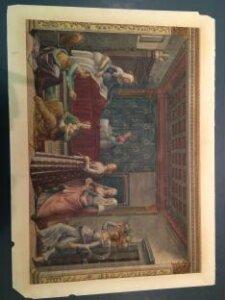}}
                
& {categories} 
& {\makecell{imagery\\ biblical\\ religion\\ copies}} 
& {\makecell{studies\\ religion}} 
& {\makecell{photograph\\ courtaulds}}
& {\makecell{paintings\\ watercolours}}   
& BAD

\\
\raisebox{-0.5\totalheight}{\includegraphics[clip, trim=0cm 1cm 0cm 1cm, width=0.15\textwidth]  {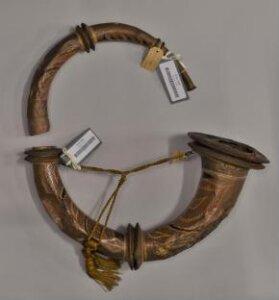}}
                
& {categories} 
& {\makecell{musical\\ objects\\ ceremonial\\ instruments}} 
& {\makecell{metalwork\\ arms}} 
& {\makecell{2003\\ slipware}}
& {\makecell{armour\\ jewellery}}   
& BAD

\\
\raisebox{-0.5\totalheight}{\includegraphics[clip, trim=0cm 1cm 0cm 1cm, width=0.15\textwidth]  {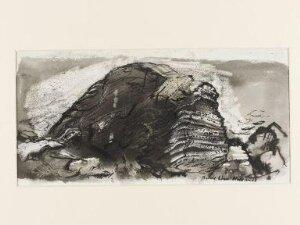}}
                
& {categories} 
& {\makecell{paintings}} 
& {\makecell{watercolours\\ paintings}} 
& {\makecell{rubbings\\ illustration}}
& {\makecell{wildlife\\ woodcuts}}   
& BAD

\\
\raisebox{-0.5\totalheight}{\includegraphics[clip, trim=0cm 3.5cm 0cm 1cm, width=0.15\textwidth]  {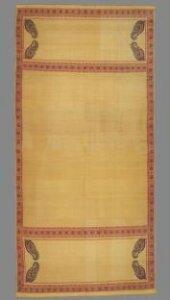}}
                
& {categories} 
& {\makecell{museum\\ india\\ textiles}} 
& {\makecell{fashion\\ it}} 
& {\makecell{board\\ india}}
& {\makecell{textiles\\ textile}}   
& BAD

\\

\end{tabular}
\end{table}

\begin{table}[]
\caption{Results of MUZE and CLIP variants compared with ground truth (GT) over the attribute: \textit{materials} }
\label{tab:muze_variants}
\scriptsize
\centering
\begin{tabular}{c|c|c|c|c|c|c|}
{Image} & Attribute                      & GT                                & MUZE                      & CLIP                      & CLIP-FA &Pred. Type \\
\midrule
\raisebox{-0.5\totalheight}{\includegraphics[clip, trim=0cm 1cm 0cm 1cm, width=0.15\textwidth]  {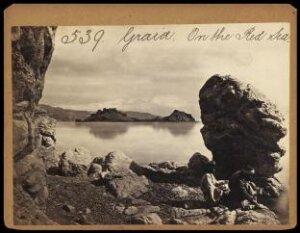}}
                
& {materials} 
& {\makecell{photographic\\ paper}} 
& {\makecell{paper\\ photographic}} 
& {\makecell{gilt-\\ courbaril}}
& {\makecell{photographic\\ colour}}   
& GOOD
\\

\raisebox{-0.5\totalheight}{\includegraphics[clip, trim=0cm 1cm 0cm 1cm, width=0.15\textwidth]  {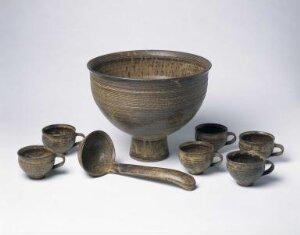}}
                
& {materials} 
& {\makecell{stoneware}} 
& {\makecell{stoneware\\ earthenware}} 
& {\makecell{earthenware\\ stoneware}}
& {\makecell{wood\\ colour}}   
& GOOD
\\

\raisebox{-0.5\totalheight}{\includegraphics[clip, trim=0cm 1cm 0cm 1cm, width=0.15\textwidth]  {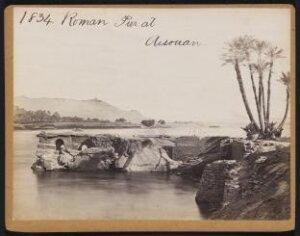}}
                
& {materials} 
& {\makecell{photographic\\ paper}} 
& {\makecell{paper\\ photographic}} 
& {\makecell{albumen\\ shisham}}
& {\makecell{photographic\\ printed}}   
& GOOD
\\
 
\raisebox{-0.5\totalheight}{\includegraphics[clip, trim=0cm 1.8cm 0cm 2cm, width=0.15\textwidth]  {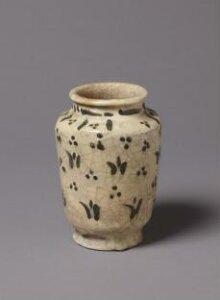}}
                
& {materials} 
& {\makecell{earthenware}} 
& {\makecell{earthenware\\ stoneware}} 
& {\makecell{stoneware\\ earthenware}}
& {\makecell{stoneware\\ earthenware}}   
& GOOD
\\
 
\raisebox{-0.5\totalheight}{\includegraphics[clip, trim=0cm 1cm 0cm 1cm, width=0.15\textwidth]  {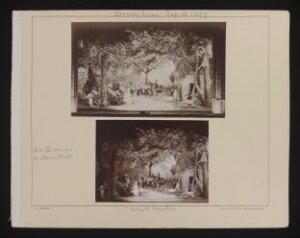}}
                
& {materials} 
& {\makecell{photographic\\ paper}} 
& {\makecell{paper\\ papier}} 
& {\makecell{albumen\\ lithographic}}
& {\makecell{printed\\ photographic}}   
& GOOD
\\

\midrule
\raisebox{-0.5\totalheight}{\includegraphics[clip, trim=0cm 1cm 0cm 2cm, width=0.15\textwidth]  {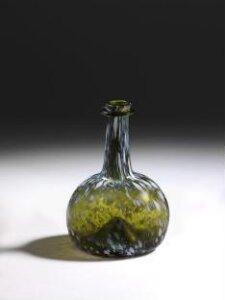}}
                
& {materials} 
& {\makecell{glass}} 
& {\makecell{unknown\\ close}} 
& {\makecell{antique\\ glaze}}
& {\makecell{glass\\ crystal}}   
& BAD
\\

\raisebox{-0.5\totalheight}{\includegraphics[clip, trim=0cm 1cm 0cm 1cm, width=0.15\textwidth]  {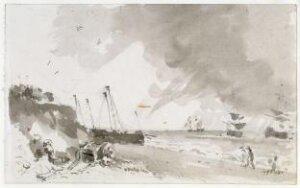}}
                
& {materials} 
& {\makecell{wash\\ fiber\\ paper\\ pencil\\ product}} 
& {\makecell{pen\\ ink}} 
& {\makecell{watercolour\\ lithographic}}
& {\makecell{paper\\ ink}}   
& BAD
\\

\raisebox{-0.5\totalheight}{\includegraphics[clip, trim=0cm 1cm 0cm 1cm, width=0.15\textwidth]  {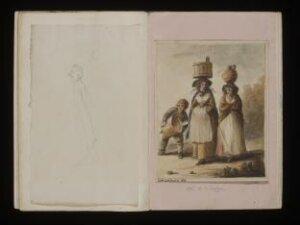}}
                
& {materials} 
& {\makecell{paper}} 
& {\makecell{watercolour\\ watercolor}} 
& {\makecell{lithographic\\ book}}
& {\makecell{paper\\ water-colour}}   
& BAD
\\

\raisebox{-0.5\totalheight}{\includegraphics[clip, trim=0cm 1cm 0cm 1cm, width=0.15\textwidth]  {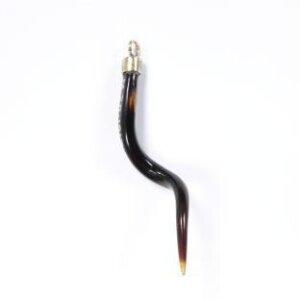}}
                
& {materials} 
& {\makecell{brass\\ horn}} 
& {\makecell{silver\\ steel}} 
& {\makecell{snake\\ horn}}
& {\makecell{colour\\ crystal}}   
& BAD
\\

\raisebox{-0.5\totalheight}{\includegraphics[clip, trim=0cm 1cm 0cm 1cm, width=0.15\textwidth]  {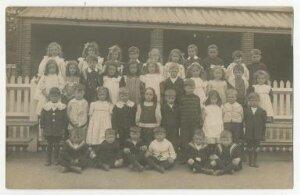}}
                
& {materials} 
& {\makecell{photographic\\ paper}} 
& {\makecell{card\\ photographic}} 
& {\makecell{unknown\\ albumen}}
& {\makecell{colour\\ printed}}   
& BAD
\\

\end{tabular}
\end{table}

\begin{table}[]
\caption{Results of MUZE and CLIP variants compared with ground truth (GT) over the attribute: \textit{techniques}  }
\label{tab:muze_variants}
\scriptsize
\centering
\begin{tabular}{c|c|c|c|c|c|c|}
{Image} & Attribute                      & GT                                & MUZE                      & CLIP                      & CLIP-FA &Pred. Type \\
\midrule
\raisebox{-0.5\totalheight}{\includegraphics[clip, trim=0cm 1cm 0cm 1cm, width=0.15\textwidth]  {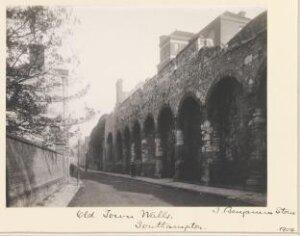}}
                
& {techniques} 
& {\makecell{pen\\ paper\\ wash}} 
& {\makecell{pen\\ ink}} 
& {\makecell{lithographic\\ book}}
& {\makecell{paper\\ pen}}   
& GOOD 
\\

\raisebox{-0.5\totalheight}{\includegraphics[clip, trim=0cm 1cm 0cm 1cm, width=0.15\textwidth]  {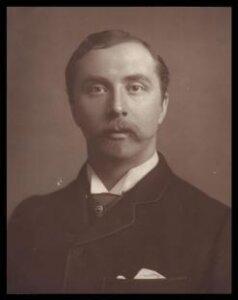}}
                
& {techniques} 
& {\makecell{photographic\\ paper}} 
& {\makecell{paper\\ photographic}} 
& {\makecell{gilt-\\ courbaril}}
& {\makecell{photographic\\ colour}}   
& GOOD 
\\
 
\raisebox{-0.5\totalheight}{\includegraphics[clip, trim=0cm 1cm 0cm 1cm, width=0.15\textwidth]  {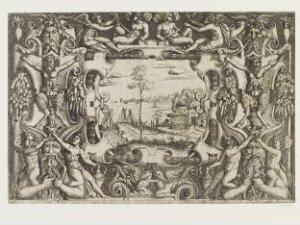}}
                
& {techniques} 
& {\makecell{stoneware}} 
& {\makecell{stoneware\\ earthenware}} 
& {\makecell{earthenware\\ stoneware}}
& {\makecell{wood\\ colour}}   
& GOOD 
\\ 
 
\raisebox{-0.5\totalheight}{\includegraphics[clip, trim=0cm 1cm 0cm 1cm, width=0.15\textwidth]  {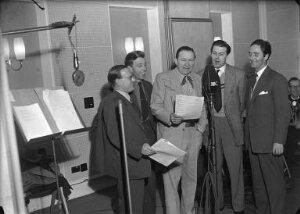}}
                
& {techniques} 
& {\makecell{photographic\\ paper}} 
& {\makecell{ paper\\ photographic}} 
& {\makecell{ albumen\\ shisham}}
& {\makecell{ photographic\\ printed}}   
& GOOD 
\\

\raisebox{-0.5\totalheight}{\includegraphics[clip, trim=0cm 1cm 0cm 1cm, width=0.15\textwidth]  {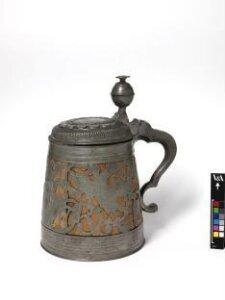}}
                
& {techniques} 
& {\makecell{photographic\\ paper}} 
& {\makecell{paper\\ papier}} 
& {\makecell{albumen\\ lithographic}}
& {\makecell{printed\\ photographic}}   
& GOOD 
\\

\midrule
\raisebox{-0.5\totalheight}{\includegraphics[clip, trim=0cm 1cm 0cm 1cm, width=0.15\textwidth]  {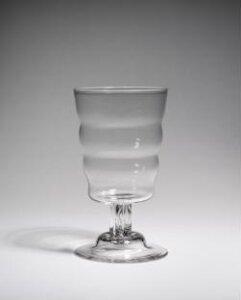}}
                
& {techniques} 
& {\makecell{earthenware}} 
& {\makecell{stoneware\\ earthenware}} 
& {\makecell{stoneware\\ earthenware}}
& {\makecell{stoneware\\ earthenware}}   
& BAD 
\\

\raisebox{-0.5\totalheight}{\includegraphics[clip, trim=0cm 1cm 0cm 1cm, width=0.15\textwidth]  {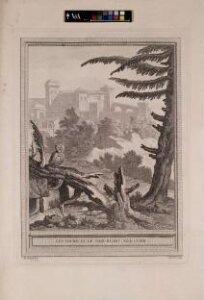}}
                
& {techniques} 
& {\makecell{paper\\ printing\\ inks}} 
& {\makecell{paper\\ papier}} 
& {\makecell{shisham\\ pashm}}
& {\makecell{colour\\ coloured}}   
& BAD 
\\

\raisebox{-0.5\totalheight}{\includegraphics[clip, trim=0cm 2cm 0cm 1cm, width=0.15\textwidth]  {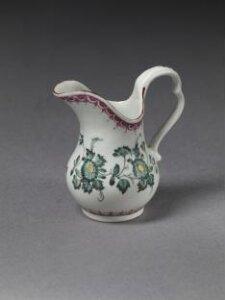}}
                
& {techniques} 
& {\makecell{earthenware\\ tin\\ glaze}} 
& {\makecell{stoneware\\ glaze}} 
& {\makecell{faience\\ jasperware}}
& {\makecell{ stoneware\\ ceramics}}   
& BAD 
\\
\raisebox{-0.5\totalheight}{\includegraphics[clip, trim=0cm 0cm 0cm 0cm, width=0.15\textwidth]  {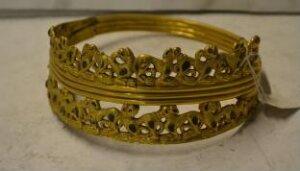}}
                
& {techniques} 
& {\makecell{plaster}} 
& {\makecell{stonepaste\\plaster}} 
& {\makecell{parian\\ india}}
& {\makecell{earthenware\\ clay}}   
& BAD 
\\

\raisebox{-0.5\totalheight}{\includegraphics[clip, trim=0cm 0cm 0cm 0cm, width=0.15\textwidth]  {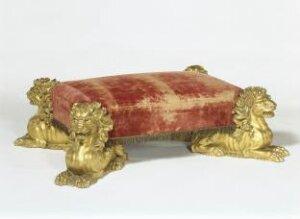}}
                
& {techniques} 
& {\makecell{paper}} 
& {\makecell{watercolour\\ watercolor}} 
& {\makecell{lithographic\\ book}}
& {\makecell{paper\\ water-colour}}   
& BAD 
\\
 
\end{tabular}
\end{table}

\begin{table}[]
\caption{Results of MUZE and CLIP variants compared with ground truth (GT) over the attribute: \textit{styles}  }
\label{tab:muze_variants}
\scriptsize
\centering
\begin{tabular}{c|c|c|c|c|c|c|}
{Image} & Attribute                      & GT                                & MUZE                      & CLIP                      & CLIP-FA &Pred. Type \\
\midrule
\raisebox{-0.5\totalheight}{\includegraphics[clip, trim=0cm 1cm 0cm 1cm, width=0.15\textwidth]  {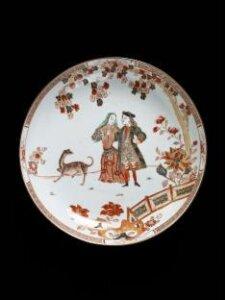}}
                
& {styles} 
& {\makecell{polychrome\\ qing\\ export\\ chinese}} 
& {\makecell{qing\\ yongzheng}} 
& {\makecell{ imari\\ kakiemon}}
& {\makecell{asia\\ chine}}   
& GOOD
\\

\raisebox{-0.5\totalheight}{\includegraphics[clip, trim=0cm 1cm 0cm 1cm, width=0.15\textwidth]  {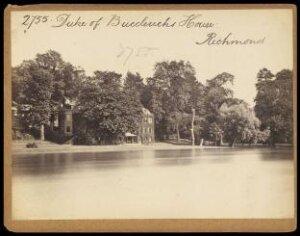}}
                
& {styles} 
& {\makecell{documentary\\ victorian}} 
& {\makecell{documentary\\ victorian}} 
& {\makecell{murshidabad\\ pictorialism}}
& {\makecell{documentary\\ unknown}}   
& GOOD
\\

\raisebox{-0.5\totalheight}{\includegraphics[clip, trim=0cm 1cm 0cm 1cm, width=0.15\textwidth]  {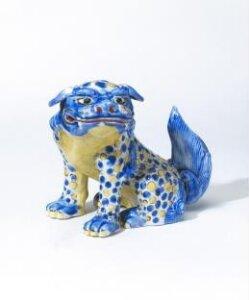}}
                
& {styles} 
& {\makecell{kakiemon\\ edo}} 
& {\makecell{edo\\ seto}} 
& {\makecell{kangxi\\ kutani}}
& {\makecell{unknown\\ mid}}   
& GOOD
\\

\raisebox{-0.5\totalheight}{\includegraphics[clip, trim=0cm 1cm 0cm 1cm, width=0.15\textwidth]  {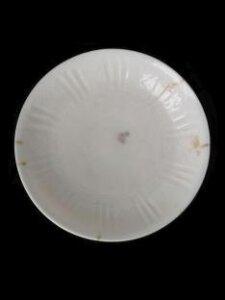}}
                
& {styles} 
& {\makecell{safavid}} 
& {\makecell{safavid\\ umayyad}} 
& {\makecell{georgian\\ kakiemon}}
& {\makecell{safavid\\ chine}}   
& GOOD
\\

\raisebox{-0.5\totalheight}{\includegraphics[clip, trim=0cm 1cm 0cm 1cm, width=0.15\textwidth]  {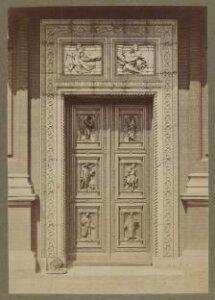}}
                
& {styles} 
& {\makecell{documentary}} 
& {\makecell{documentary\\ photojournalism}} 
& {\makecell{jugendstil\\ palladian}}
& {\makecell{unknown\\ arts}}   
& GOOD
\\

\midrule
\raisebox{-0.5\totalheight}{\includegraphics[clip, trim=0cm 1cm 0cm 1cm, width=0.15\textwidth]  {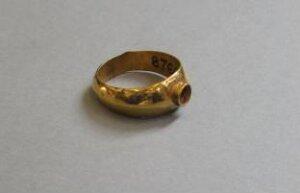}}

& {styles} 
& {\makecell{roman\\ style\\ or\\ period}} 
& {\makecell{unknown\\ urban}} 
& {\makecell{golden\\ antique}}
& {\makecell{roman\\ greek}}   
& BAD
\\

\raisebox{-0.5\totalheight}{\includegraphics[clip, trim=0cm 1cm 0cm 1cm, width=0.15\textwidth]  {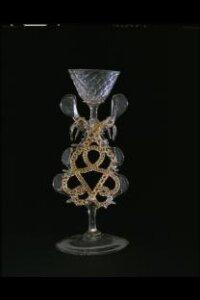}}

& {styles} 
& {\makecell{venise\\ facon\\ de\\ europe}} 
& {\makecell{european\\ europe}} 
& {\makecell{judaica\\ verre}}
& {\makecell{italy\\ europe}}   
& GBAD
\\

\raisebox{-0.5\totalheight}{\includegraphics[clip, trim=0cm 1cm 0cm 1cm, width=0.15\textwidth]  {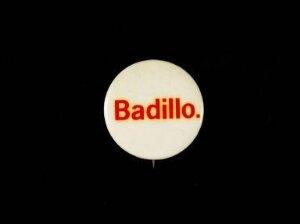}}

& {styles} 
& {\makecell{half\\ 20th\\ second\\ century}} 
& {\makecell{unknown\\ urban}} 
& {\makecell{basohli\\ bauhaus}}
& {\makecell{unknown\\ mid}}   
& BAD
\\

\raisebox{-0.5\totalheight}{\includegraphics[clip, trim=0cm 3cm 0cm 3cm, width=0.15\textwidth]  {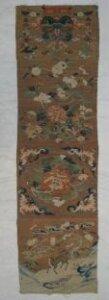}}

& {styles} 
& {\makecell{kangxi}} 
& {\makecell{qing\\ ming}} 
& {\makecell{antique\\ kutani}}
& {\makecell{qing\\ sui-tang}}   
& BAD
\\
\raisebox{-0.5\totalheight}{\includegraphics[clip, trim=0cm 1cm 0cm 1cm, width=0.15\textwidth]  {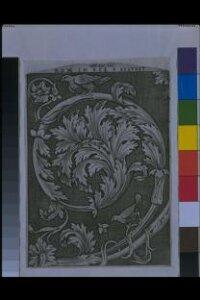}}

& {styles} 
& {\makecell{renaissance}} 
& {\makecell{unknown\\ urban}} 
& {\makecell{jacobean\\ jugendstil}}
& {\makecell{ renaissance\\ spring}}   
& BAD
\\

\end{tabular}
\end{table}